\documentclass[10pt,journal,compsoc]{IEEEtran}
\usepackage{ifpdf}
\ifpdf
\else
\fi
\ifCLASSOPTIONcompsoc
  \usepackage[nocompress]{cite}
\else
  \usepackage{cite}
\fi
\ifCLASSINFOpdf
\else
\fi
\ifCLASSINFOpdf
 \usepackage[pdftex]{thumbpdf}
\else
 \usepackage[dvips]{thumbpdf}
\fi
\newcommand\MYhyperrefoptions{bookmarks=true,bookmarksnumbered=true,
pdfpagemode={UseOutlines},plainpages=false,pdfpagelabels=true,
colorlinks=true,linkcolor={black},citecolor={black},urlcolor={black},
pdftitle={Bare Demo of IEEEtran.cls for Computer Society Journals},%<!CHANGE!
pdfsubject={Typesetting},%<!CHANGE!
pdfauthor={Michael D. Shell},%<!CHANGE!
pdfkeywords={Computer Society, IEEEtran, journal, LaTeX, paper,
             template}}%<^!CHANGE!
\ifCLASSINFOpdf
\usepackage[\MYhyperrefoptions,pdftex]{hyperref}
\else
\usepackage[\MYhyperrefoptions,breaklinks=true,dvips]{hyperref}
\usepackage{breakurl}
\fi
\usepackage{graphicx}
\usepackage{mathtools}
\usepackage{color}
\usepackage{algorithm}
\usepackage{algpseudocode}
\usepackage{amsmath}
\usepackage{setspace}
\usepackage{multirow}
\usepackage{booktabs}
\usepackage{makecell}
\usepackage{amsfonts}
\usepackage{wrapfig}
\usepackage{graphicx}
\usepackage{subfigure}
\usepackage{kotex}
\usepackage[switch]{lineno}  % line number
\usepackage{xcolor}
\usepackage{pifont}
\usepackage{fontawesome}
\usepackage{hyperref}
\usepackage{colortbl}
\usepackage{soul} %added by 선영 for \st{} cancel midline
\usepackage{booktabs}
\usepackage{threeparttable}
\usepackage{tabularx}

\newcommand{\hjk}[1]{\textcolor{black}{#1}}
\newcommand{\dasol}[1]{\textcolor{black}{#1}}
\DeclareMathOperator*{\argmin}{\arg\!\min}

\def\wbst{\mathbf{w}^{\ast}}
\newcommand{\Vc}{\mathcal{V}}

\newcommand{\omitme}[1]{}
\newcommand{\Lc}{\mathcal{L}}
\newcommand{\Tc}{\mathcal{T}}

\newcommand{\Rb}{\mathbb{R}}

\def\ie{\emph{i.e}.}

\def\wb{\mathbf{w}}

\def\wb{\mathbf{w}}

\def\Zb{\mathbf{Z}}

\def\Rb{\textbf{R}}

\newcommand{\EE}{\mathbb{E}}

\makeatletter
\newcount\my@repeat@count
\newcommand{\myrepeat}[2]{%
  \begingroup
  \my@repeat@count=\z@
  \@whilenum\my@repeat@count<#1\do{#2\advance\my@repeat@count\@ne}%
  \endgroup
}
\makeatother
\begin{document}

\title{Self-supervised Auxiliary Learning \\for Graph Neural Networks via Meta-Learning}


\author{Dasol~Hwang,
        Jinyoung~Park,
        Sunyoung~Kwon,
        Kyung-Min~Kim,
        Jung-Woo~Ha,
        and Hyunwoo~J.~Kim% <-this % stops a space
\IEEEcompsocitemizethanks{\IEEEcompsocthanksitem D. Hwang, J. Park and H. J. Kim are with the Department of Computer Science and Engineering, Korea University, Seoul 02841, Korea.\protect\\
E-mail: \{dd\_sol, lpmn678, hyunwoojkim\}@korea.ac.kr}
\IEEEcompsocitemizethanks{\IEEEcompsocthanksitem S. Kwon is with School of Biomedical Convergence Engineering, College of Information and Biomedical Engineering, Pusan National University, Yangsan 50612, Korea. \protect\\
E-mail: skwon@pusan.ac.kr}
\IEEEcompsocitemizethanks{\IEEEcompsocthanksitem K. Kim and J. Ha are with NAVER AI LAB and NAVER CLOVA, NAVER Corp., Bundang 13561, South Korea.\protect\\
E-mail: \{kyungmin.kim.ml, jungwoo.ha\}@navercorp.com}
\IEEEcompsocitemizethanks{\IEEEcompsocthanksitem First two authors have equal contribution.}
\IEEEcompsocitemizethanks{\IEEEcompsocthanksitem H. J. Kim is the corresponding author.}
}

\IEEEtitleabstractindextext{%
\begin{abstract}
In recent years, graph neural networks (GNNs) have been widely adopted in the representation learning of graph-structured data and provided state-of-the-art performance in various applications such as link prediction, node classification, and recommendation.
Motivated by recent advances of self-supervision for representation learning in natural language processing and computer vision, self-supervised learning has been recently studied to leverage unlabeled graph-structured data. 
However, employing self-supervision tasks as auxiliary tasks to assist a primary task has been less explored in the literature on graphs.
In this paper, we propose a novel self-supervised auxiliary learning framework to effectively learn graph neural networks.
Moreover, this work is the first study showing that a meta-path prediction is beneficial as a self-supervised auxiliary task for heterogeneous graphs.
Our method is learning to learn a primary task with various auxiliary tasks to improve generalization performance. 
The proposed method identifies an effective combination of auxiliary tasks and automatically balances them to improve the primary task. 
Our methods can be applied to any graph neural network in a plug-in manner without manual labeling or additional data. 
Also, it can be extended to any other auxiliary tasks.
Our experiments demonstrate that the proposed method consistently improves the performance of node classification and link prediction.

\end{abstract}
% Note that keywords are not normally used for peer review papers.
}
\maketitle

% \IEEEtitleabstractindextext{%
% \begin{abstract}
% The abstract goes here.
% \end{abstract}

% % Note that keywords are not normally used for peerreview papers.
% \begin{IEEEkeywords}
% Computer Society, IEEE, IEEEtran, journal, \LaTeX, paper, template.
% \end{IEEEkeywords}}

\maketitle

% is not selected <OR> if conference mode is selected - because compsoc
% conference papers position the abstract like regular (non-compsoc)
\IEEEdisplaynontitleabstractindextext

% \fi
\IEEEpeerreviewmaketitle

\IEEEraisesectionheading{\section{Introduction}}\label{sec:introduction}
\IEEEPARstart{R}{ecently}, there has been increasing research interest in Graph Neural Networks (GNNs) for effective representation learning on graph-structured data.
GNNs~\cite{hamilton2017representation, wu2020comprehensive, zhou2018graph} have shown superior performance in a variety of tasks such as node classification~\cite{GCN}, link prediction~\cite{link}, and graph classification~\cite{DiffPool}.
The representation learned by GNNs yields state-of-the-art performance in a wide range of applications including social network analysis~\cite{social1}, citation network analysis~\cite{GCN}, recommender systems~\cite{recom1, wang2018ripplenet, wang2019knowledge}, physics~\cite{battaglia2016interaction}, and drug discovery~\cite{wu2018moleculenet}, visual understanding~\cite{yang2018graph,HwangRTKCS18}.
% With the success of GNNs~\cite{wu2020comprehensive, zhou2018graph}, the need has emerged for better representation learning on graphs.
As the success of GNNs raises the need for better representation learning in graphs, most modern deep learning approaches have been extended to graphs, taking into account the characteristics of graphs. 
These include pre-training and fine-tuning~\cite{devlin2018bert}, transfer learning~\cite{pan2009survey}, multi-task learning~\cite{auxlearning1}, self-supervised learning~\cite{you2020does}, and meta-learning~\cite{MAML}.
However, learning with auxiliary (pre-text) tasks has been less explored for further improving generalization performance on graphs.

Learning with auxiliary tasks is to jointly learn representation for multiple tasks by sharing the representation between them, especially to aid in a primary task.
However, there are few approaches to generalize this method to graph-structured data due to their fundamental challenges.
First, graph structure (e.g., the number of nodes/edges, and diameter) and its meaning can significantly differ between domains.
So, the model trained on auxiliary tasks can harm generalization on the primary task, \ie, \textit{negative transfer}~\cite{pan2009survey}. It means that auxiliary tasks can dominate training and degrade the performance of the primary task.
Also, many graph neural networks are transductive approaches. This often makes knowledge transfer between datasets inherently infeasible.
So, these assume that the auxiliary tasks are carefully selected with substantial domain knowledge and expertise in graph characteristics.
For this, we propose a novel auxiliary learning method to identify an effective combination of auxiliary tasks and automatically balances them to improve the primary task.

Recent works show that the representation can be further improved by self-supervised learning on graphs.
Self-supervised tasks on graphs can be designed based on node attributes and graph topology: node property (e.g., degree, centrality, clustering coefficient, PageRank score etc.) prediction~\cite{jin2020self}, context prediction~\cite{hu2020strategies}, graph partitioning~\cite{you2020does}, cluster detection~\cite{huunsupervised}, graph kernel~\cite{navarin2018pre}, graph completion~\cite{you2020does}, and attribute masking~\cite{hu2020strategies}. 
These tasks can be utilized as auxiliary tasks.
However, since most graph neural networks operate on \textit{homogeneous} graphs, which have a single type of nodes and edges, these auxiliary tasks are not specifically designed for \textit{heterogeneous graphs}, which have multiple types of nodes and edges. 
Heterogeneous graphs commonly occur in real-world applications, for instance, a music dataset has multiple types of nodes (e.g., user, song, artist) and multiple types of relations (e.g., user-artist, song-film, song-instrument).
Thus, we introduce meta-path prediction as a self-supervised auxiliary task to leverage rich information of heterogeneous graphs.

In this paper, we propose a framework to train graph neural networks with automatically selected auxiliary self-supervised tasks which assist the primary task without manual labeling or additional data. This can be formulated as a meta-learning problem. Our approach first generates meta-paths from heterogeneous graphs without manual labeling and train a model with meta-path prediction (auxiliary task) to assist the primary task such as link prediction and node classification. Furthermore, it can be extended to any other auxiliary tasks such as supervised and self-supervised tasks. Our method can be adopted to existing GNNs in a plug-in manner, enhancing the model performance.

Our \textbf{contributions} are as follows: \textbf{(i)} We propose a self-supervised learning method on a heterogeneous graph via meta-path prediction without manual labeling or additional data. 
\textbf{(ii)} Our framework automatically selects auxiliary tasks (meta-paths) to assist the primary task via meta-learning. 
\textbf{(iii)} We develop Hint Network that helps the learner network to benefit from challenging auxiliary tasks. 
To the best of our knowledge, this is the first auxiliary task with meta-paths specifically designed for leveraging heterogeneous graph structure.
Our experiments show that various auxiliary tasks improve the representational power and meta-path prediction provides a significant improvement as the auxiliary task.
Moreover, we confirm that the gains can be further improved by our method that explicitly optimizes auxiliary tasks for the primary task via meta-learning and Hint network.

\section{Related Work}
\subsection{Graph Neural Networks} 
Graph neural networks (GNNs) have been proposed to learn the representations of graph-structured data. It was developed by extending the existing deep learning approaches to graph data. 
Motivated by the success of CNNs, Kipf \& Welling~\cite{GCN} presented a popular graph neural network called GCN, which bridges the gap between spectral approaches and spatial approaches by localizing first-order approximation of graph convolutions.   
Graph attention network (GAT)~\cite{GAT} which adopts the attention mechanism to learn the attention coefficient of connected node pair is proposed.
\cite{xu2018powerful} demonstrated that the existing message passing method can not distinguish structurally different graphs. Then it proposed a graph isomorphism network (GIN) that can complement it.
SGC~\cite{wu2019simplifying} is a simplified model by removing non-linearity between GCN layers.
GTN~\cite{yun2019graph} can learn new graph structures on heterogeneous graphs by softly selecting edge types and composite relations for generating useful multi-hop connections.
Based on the variants of GNNs, they have provided promising results in a wide range of applications on graphs.

\subsection{Self-supervised Learning} 
Self-supervised learning is a novel learning framework to train deep learning models by assigning supervision to unlabeled data by itself.
These methods are constantly being explored to effectively utilize rich unlabeled data.
In computer vision, various self-supervised learning methods have been developed for general semantic understanding of images. 
Relative positioning~\cite{ss1} aims to train networks to predict the relative position of two regions in the region patches of an image. 
\cite{ss2} was inspired by Jigsaw Puzzle and designed a pretext task to train a model to solve the puzzle.
Following this, various self-supervised learning methods have been proposed: rotation~\cite{gidaris2018unsupervised}, colourization~\cite{zhang2016colorful}, exemplar~\cite{dosovitskiy2014discriminative},
image inpainting~\cite{pathak2016context} and image clustering~\cite{caron2018deep}.

Inspired by these efforts, recent works show promising results that self-supervised learning can be effective for GNNs. 
Hu et al.~\cite{hu2020strategies} has introduced several strategies for pre-training GNNs in the graph-level tasks, such as attribute masking, context prediction, and structural similarity prediction.
\cite{you2020does} incorporates self-supervision into GCNs and introduced three novel self-supervised tasks (node clustering, graph partitioning, graph completion) for GCNs on multi-task learning.
These self-supervised tasks can be utilized as auxiliary tasks.
However, all auxiliary tasks may not be beneficial for the primary task.
Thus, we studied the \textit{auxiliary learning} method for GNNs that explicitly focus on the primary task.

\subsection{Learning with Auxiliary tasks} % Auxiliary Learning
Multi-task learning (MTL) \cite{zhang2017survey} is usually used to improve the performance of all tasks by utilizing useful information from multiple relevant tasks. 
However, even with relevant tasks, MTL often suffers from degradation due to the competition between tasks.
Therefore, learning with multiple auxiliary tasks can harm generalization on the primary task referred to as negative transfer.
To address this, Auxiliary Learning has been proposed \cite{zhang2014aux}.
Auxiliary Learning is a learning strategy to employ auxiliary tasks to assist the primary task. 
It is similar to multi-task learning, but auxiliary learning cares only about the performance of the primary task.
These methods are proposed in a wide range of tasks~\cite{auxlearning1, Unaux1, Unaux2}. 
AC-GAN~\cite{acgan} proposed an auxiliary classifier for generative models.
In this paper, we proposed a self-supervised auxiliary learning framework that generates meta-paths on heterogeneous graphs to make new labels and trains models to predict meta-paths as auxiliary tasks.

\subsection{Meta-learning} 
Meta-learning aims at learning to learn models efficiently and effectively, and generalizes the learning strategy to new tasks. It has been proved to be effective on few-shot learning.
Meta-learning includes black-box methods to approximate gradients without any information about models~\cite{santoro2016meta, DBLP:conf/iclr/RaviL17},  
optimization-based methods to learn an optimal initialization for adapting new tasks~\cite{MAML, han2018coteaching, layerwise}, 
learning loss functions~\cite{han2018coteaching,learnwhere} and  
metric-learning or non-parametric methods for few-shot learning \cite{Koch2015SiameseNN, DBLP:conf/nips/SnellSZ17, DBLP:conf/cvpr/SungYZXTH18}.
In contrast to classical learning algorithms that generalize across samples, meta-learning generalizes across tasks.
In contrast to classical learning algorithms that generalize the concept across samples, meta-learning generalizes across tasks.
In this paper, we use meta-learning to learn a concept across tasks and transfer the knowledge from auxiliary tasks to the primary task.

Recent works on meta-learning for graphs have emerged.
\cite{xiong2018one} is one-shot relational learning framework to learn a differentiable metric to match entity pairs. 
Meta-GNN~\cite{zhou2019meta} and Meta-Graph~\cite{bose2019meta} are gradient-base meta-learning approaches in few-shot setting for node classification and link prediction.
MetaR~\cite{chen2019meta} is a meta relational learning framework to focus on transferring relation-specific meta information in few-shot link prediction.
Besides, G-META~\cite{huang2020graph} can learn transferable knowledge faster via meta gradients by leveraging local subgraphs. 
GPN~\cite{liu2019learning} is a novel meta-learner that explicitly relates tasks on a graph explaining the relations of predicted classes in few-shot learning.
In this work, via meta-learning, we propose a novel method that learns how to assist the primary task with auxiliary tasks by optimizing a weighting function.
\begin{figure*}[!htbp]
  \includegraphics[scale=0.56, bb=0 0 0 250]{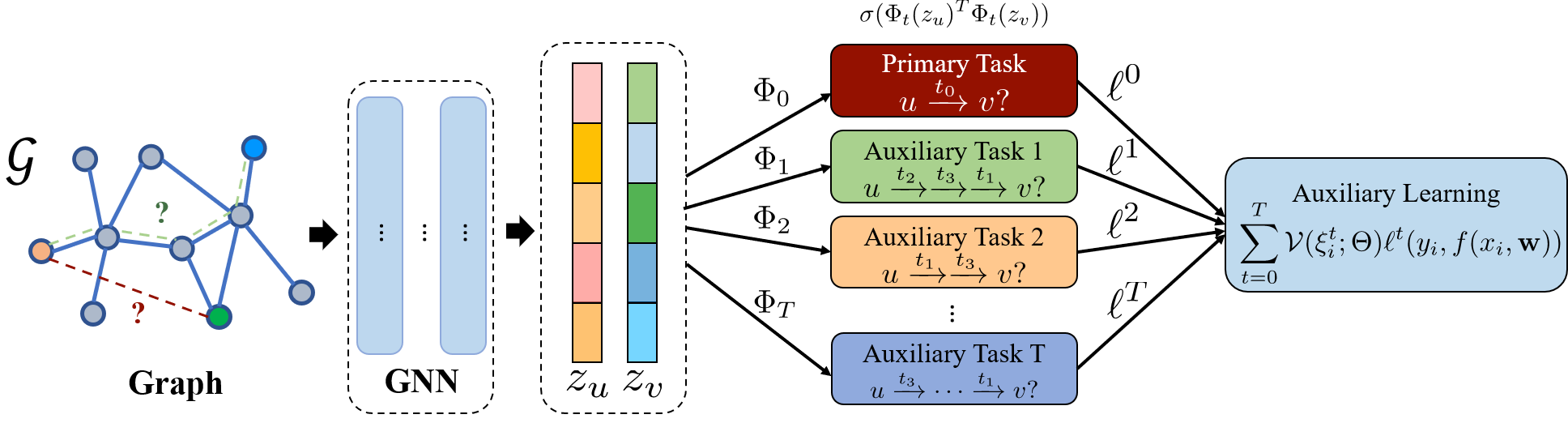}
  \caption{The SELAR framework for self-supervised auxiliary learning. Our framework learns how to balance (or softly select) auxiliary tasks to improve the primary task via meta-learning. In this paper, the primary task is link prediction (or node classification) and auxiliary tasks are meta-path predictions to capture rich information of a heterogeneous graph.}
  \label{fig:meta-path-prediction}
\end{figure*}
\section{Method}
\label{headings}
The goal of our framework is to learn with multiple auxiliary tasks to improve the performance of the primary task. 
In this work, we demonstrate the benefits of our framework with various auxiliary tasks.
\dasol{Especially, we evaluate meta-path prediction as auxiliary tasks.} 
The meta-paths capture diverse and meaningful relations between nodes on heterogeneous graphs~\cite{HAN}.
However, learning with auxiliary tasks for graphs has multiple challenges: identifying useful auxiliary tasks, balancing the auxiliary tasks with the primary task, and converting challenging auxiliary tasks into solvable (and relevant) tasks. 
To address the challenges, we propose \textbf{SEL}f-supervised \textbf{A}uxiliary Lea\textbf{R}ning (\textbf{SELAR}). 

\dasol{Our framework consists of two main components: 1) We propose a novel framework that learns weight functions to softly select auxiliary tasks and balance them with the primary task via meta-learning. 2) We introduce Hint Networks to convert challenging auxiliary tasks into more relevant and solvable tasks to the primary task learner.}
\subsection{Meta-path Prediction as a self-supervised task}
Most existing graph neural networks have been studied focusing on {\em homogeneous graphs} that have a single type of nodes and edges.
However, in real-world applications, {\em heterogeneous graphs}~\cite{heterogeneous}, which have multiple types of nodes and edges, commonly occur. 
Learning models on the heterogeneous graphs requires different considerations to effectively represent their node and edge heterogeneity. 

\vspace{0.3cm}
{\textbf{Heterogeneous graph.} } 
Let $G = (V,E)$ be a graph with a set of nodes $V$ and edges $E$. 
A heterogeneous graph~\cite{shi2016survey} is a graph equipped with a node type mapping function $f_v : V \rightarrow \mathcal{T}^v$ and an edge type mapping function  $f_e : E \rightarrow \mathcal{T}^e$, where  $\mathcal{T}^v$ is a set of node types and $\mathcal{T}^e$ is a set of edge types. 
Each node $v_i \in V$ (and edge $e_{ij} \in E$ resp.) has one node type, i.e., $f_v(v_i) \in \mathcal{T}^v$, (and one edge type $f_e(e_{ij}) \in  \mathcal{T}^e$ resp.). In this paper, we consider the heterogeneous graphs with $|\mathcal{T}^e|> 1$ or $|\mathcal{T}^v|> 1$. When $|\mathcal{T}^e|=1$ and $|\mathcal{T}^v|=1$, it becomes a homogeneous graph.

\vspace{0.3cm}
{\textbf{Meta-Path.} } 
Meta-path~\cite{HAN,sun2011pathsim}  is a path on a heterogeneous graph $G$ that a sequence of nodes connected with heterogeneous edges, i.e., ${v}_1 \xrightarrow{t_1} {v}_2 \xrightarrow{t_2} \ldots  \xrightarrow{t_l} {v}_{l+1}$, 
where $t_l \in \mathcal{T}^e$ denotes an $l$-th edge type of the meta-path. 
The meta-path can be viewed as a composite relation $R = t_1 \circ t_2 \ldots \circ t_l$ between node ${v}_1$ and ${v}_{l+1}$, where $R_1 \circ R_2$ denotes the composition of relation $R_1$ and $R_2$.
The definition of meta-path generalizes multi-hop connections and is shown to be useful to analyze heterogeneous graphs.
For instance, in Book-Crossing dataset, `user-item-written.series-item-user' indicates that a meta-path that connects users who like the same book series.

\vspace{0.3cm}
We introduce \textbf{meta-path prediction} as a self-supervised auxiliary task to improve the representational power of graph neural networks. 
To our knowledge, the meta-path prediction has not been studied in the context of self-supervised learning for graph neural networks in the literature.

{\textbf{Meta-path prediction.} } 
Meta-path prediction is similar to link prediction but meta-paths allow heterogeneous composite relations. 
The meta-path prediction can be achieved in the same manner as link prediction. 
If two nodes $u$ and $v$ are connected by a meta-path $p$ with the heterogeneous edges $(t_1, t_2,\ldots t_\ell)$,  then $y_{u,v}^p=1$, otherwise $y_{u,v}^p=0$. The labels can be generated from a heterogeneous graph without any manual labeling. 
They can be obtained by $A_p = A_{t_l} \ldots  A_{t_2}A_{t_1}$, where $A_{t}$ is the adjacency matrix of edge type $t$. The binarized value at $(u, v)$ in $A_p$ indicates whether  $u$ and $v$ are connected with the meta-path $p$.
In this paper, we use meta-path prediction as a self-supervised auxiliary task. 

Let $\mathbf{X} \in \Rb^{|V| \times d }$ and $\mathbf{Z} \in \Rb^{|V| \times d'}$ be input features and their hidden representations learnt by GNN $f$, i.e., $\Zb = f(\mathbf{X};\mathbf{w},\mathbf{A})$, where $\mathbf{w}$ is the parameter for $f$, and $\mathbf{A} \in \Rb^{|V|\times |V|}$ is the adjacency matrix. Then link prediction and meta-path prediction are obtained by a simple operation as
\begin{align}
\hat{y}_{u,v}^t = \sigma (\Phi_t(z_u)^\top \Phi_t(z_v) ),
\end{align}
where $\Phi_t$ is the task-specific network for task $t \in \Tc$ and $z_u$ and $z_v$ are the node embeddings of node $u$ and $v$. e.g., $\Phi_0$ (and $\Phi_1$ resp.) for link prediction (and the first type of meta-path prediction resp.).

\vspace{0.3cm}
\textbf{The architecture} is shown in Fig.~\ref{fig:meta-path-prediction} \dasol{for link prediction, especially meta-path prediction as auxiliary tasks.}
To optimize the model, as the link prediction and the meta-path prediction, cross entropy is used.
The graph neural network $f$ is shared by the link prediction and meta-path predictions \dasol{and there is a task-specific function $\Phi$ for each task.}
As any auxiliary learning methods, the auxiliary task samples (meta-paths) should be carefully chosen and properly weighted so that the auxiliary task (meta-path prediction) does not compete with link prediction (primary task) especially when the capacity of GNNs is limited.
To address these issues, we propose our framework that automatically selects meta-paths and balances them with the primary task via meta-learning. 

\label{meta-path-prediction}
\subsection{Self-Supervised Auxiliary Learning}
Our framework SELAR is learning to learn a primary task with multiple auxiliary tasks to assist the primary task.
This can be formally written as 
\begin{align}
&\min_{\wb,\Theta} \;\; \underset{(x,y) \sim D^{pr}\myrepeat{30}{\;}}{\text{\large $\EE$} \;\; \left [\; \Lc^{pr}(\wb^{\ast}(\Theta)) \; \right ]}  \nonumber  \\
&\text{ s.t. }\;\; \wb^{\ast}(\Theta) = \argmin_{\wb} \underset{(x,y) \sim D^{pr+au}\myrepeat{25}{\;}} {\;\; \EE \;\;\left [\;\Lc^{pr+au}(\wb;\Theta) \; \right ]},
\label{eq:probsetup}
\end{align}

where $\Lc^{pr}(\cdot)$ is the primary task loss function to evaluate the trained model $f(x; \wb^{\ast}(\Theta))$ on meta-data (a validation for meta-learning~\cite{han2018coteaching}) $D^{pr}$ and $\Lc^{pr+au}$ is the loss function to train a model on training data $D^{pr+au}$ with the primary and auxiliary tasks. To avoid cluttered notation, $f$, $x$, and $y$ are omitted. Each task $\mathcal{T}_t$ has $N_t$ samples and $\Tc_0$ and $\{\Tc_t\}_{t=1}^T$ denote the primary and auxiliary tasks respectively. 

The proposed formulation in Eq.~\eqref{eq:probsetup} learns how to assist the primary task by optimizing $\Theta$ via meta-learning. The nested optimization problem given $\Theta$ is a regular training with properly adjusted loss functions to balance the primary and auxiliary tasks. The formulation can be more specifically written as 
\begin{align}
\min_{\wb, \Theta} & \sum_{i=1}^{M} \frac{1}{M}  \ell^0(y_i^{(0,meta)}, f(x_i^{(0,meta)};\wbst(\Theta))) \label{eq:Lmeta} \\ 
\text{s.t. } & \wbst(\Theta) = \argmin_{\wb} \sum_{t = 0}^T \sum_{i=1}^{N_t} \frac{1}{N_t} \Vc(\cdot;\Theta) \ell^t(y_i, \hat{y}_i) \label{eq:Ltrain},
\end{align}
where $\ell^t$ and $f^t$ denote the loss function and the model for task $t$. We overload $\ell^t$ with its function value, i.e., $\ell^t = \ell^t(y_i, \hat{y}_i)$ where $\hat{y}_i = f^t(x_i^{(t,train)};\wb)$.
The embedding vector of $i_{th}$ sample for task $t$ denotes $\xi^{(t, train)}_i$. The input of the weighting function is $\xi^t_i$. i.e., $\Vc(\xi^t_i; \Theta)$. 
In our experiment, $\xi^t_i$ is the concatenation of the loss value, one-hot representation of task types, and the label of the sample (positive/negative), i.e.,  $\xi^t_i = \left [\ell^t; e_t; y_i^{(t,train)} \right ] \in \Rb^{T+2}$.

\begin{algorithm}
\caption{\label{alg:selar}Self-supervised Auxiliary Learning} 
\textbf{Input:}  training data for primary/auxiliary tasks $D^{pr}/D^{au}$,  mini-batch size $N_{pr}, N_{au}$ \newline
\textbf{Input:}  max iterations $K$, \# folds for cross validation $C$ \newline
\textbf{Output:}  network parameter $\wb^K$ for the primary task 
% \setstretch{1.3}
\setstretch{1.5}
\begin{algorithmic}[1]
\For {$k=1$ to $K$}
\State $D^{pr}_{m} \leftarrow \text{MiniBatchSampler}(D^{pr}, N_{pr})$ 
\State $D^{au}_{m} \leftarrow \text{MiniBatchSampler}(D^{au}, N_{au})$ 
    \For {$c=1$ to $C$} \Comment{Cross Validation}
    \State  $D^{pr(train)}_{m}, D^{pr(meta)}_{m} \leftarrow \text{CVSplit}(D^{pr}_m, c)$ 
    \State  
    $\hat{\wb}^k(\Theta) \leftarrow \wb^k - \alpha \nabla_{\wb} \Lc^{pr+au}(\wb^k; \Theta)$ \\
    \;\;\;\;\;\;\;\;\;\;  with $D^{pr(train)}_{m} \cup D^{au}_m$ \Comment{Eq. \eqref{eq:w_hat}}
    \State $g_c \leftarrow \nabla_{\Theta}\Lc^{pr}(\hat{\wb}^k)$ with $D^{pr(meta)}_{m}$ \Comment{Eq. \eqref{eq:updateTheta}}
    \EndFor
\State Update $\Theta^{k+1} \leftarrow \Theta^{k} - \beta \sum_{c}^C g_c$ \Comment{Eq.\eqref{eq:updateThetaCV}} 
\State $\wb^{k+1} = \wb^{k} - \alpha \nabla_{\wb}\Lc^{pr+au} (\wb^k;\Theta^{k+1})$ \\
\;\;\;\;\;\; with $D^{pr}_m \cup D^{au}_m$ \Comment{Eq. \eqref{eq:updateW}}
\EndFor
\end{algorithmic} 
\end{algorithm}

% \begin{algorithm}
% \caption{\label{alg:selar}Self-supervised Auxiliary Learning} 
% \textbf{Input:}  training data for primary/auxiliary tasks $D^{pr}, D^{au}$,  mini-batch size $N_{pr}, N_{au}$ \newline
% \textbf{Input:}  max iterations $K$, \# folds for cross validation $C$, learning rate $\alpha, \beta$ \newline
% \textbf{Output:}  network parameter $\wb^K$ for the primary task 
% \setstretch{1.3}
% %\setstretch{1.35}
% \begin{algorithmic}[1]
% \State Initialize $\wb^1, \Theta^1$ 
% \For {$k=1$ to $K$}
% \State $D^{pr}_{m} \leftarrow \text{MiniBatchSampler}(D^{pr}, N_{pr})$ 
% \State $D^{au}_{m} \leftarrow \text{MiniBatchSampler}(D^{au}, N_{au})$ 
%     \For {$c=1$ to $C$} \Comment{Meta Learning with Cross Validation}
%     \State  $D^{pr(train)}_{m}, D^{pr(meta)}_{m} \leftarrow \text{CVSplit}(D^{pr}_m, c)$ 
%     \Comment{Split Data for CV}
%     \State  
%     $\hat{\wb}^k(\Theta^k) \leftarrow \wb^k - \alpha \nabla_{\wb} \Lc^{pr+au}(\wb^k; \Theta^k)$ with $D^{pr(train)}_{m} \cup D^{au}_m$ \Comment{Eq. \eqref{eq:w_hat}}
%     \State $g_c \leftarrow \nabla_{\Theta}\Lc^{pr}(\hat{\wb}^k(\Theta^k))$ with $D^{pr(meta)}_{m}$ \Comment{Eq. \eqref{eq:updateTheta}}
%     \EndFor
% \State Update $\Theta^{k+1} \leftarrow \Theta^{k} - \beta \sum_{c}^C g_c$ \Comment{Eq. \eqref{eq:updateThetaCV}} 
% \State $\wb^{k+1} = \wb^{k} - \alpha \nabla_{\wb}\Lc^{pr+au} (\wb^k;\Theta^{k+1})$ with $D^{pr}_m \cup D^{au}_m$ \Comment{Eq. \eqref{eq:updateW}}
% \EndFor
% \end{algorithmic} 
% \end{algorithm}

To derive our learning algorithm, 
we first shorten the objective function in Eq.~\eqref{eq:Lmeta} and Eq.~\eqref{eq:Ltrain} as $\Lc^{pr}(\wbst(\Theta))$ and $\Lc^{pr+au}(\wb; \Theta)$.
This is equivalent to Eq.~\eqref{eq:probsetup} without expectation.
Then, our formulation is given as 
\begin{equation}
\min_{\wb,\Theta}  \Lc^{pr}(\wbst(\Theta)) \;\; \text{ s.t. } \wbst(\Theta) = \argmin_{\wb} \Lc^{pr+au}(\wb; \Theta),
\label{eq:shortformulation}
\end{equation}
To circumvent the difficulty of the bi-level optimization, as previous works~\cite{MAML, han2018coteaching} in meta-learning we approximate it with the updated parameters $\hat{\wb}$ using the gradient descent update as
\begin{align}
\wbst(\Theta) \approx \hat{\wb}^k(\Theta^k) = \wb^k - \alpha \nabla_{\wb} \Lc^{pr+au}(\wb^k; \Theta^k), \label{eq:w_hat}
\end{align}
where $\alpha$ is the learning rate for $\wb$. 
We do not numerically evaluate $\hat{\wb}^k(\Theta)$ instead we plug the computational graph of $\hat{\wb}^k$ in  $\Lc^{pr}(\wbst(\Theta))$ to optimize $\Theta$. 
Let $\nabla_{\Theta} \Lc^{pr}(\wbst(\Theta^k))$ be the gradient evaluated at $\Theta^k$.
Then updating parameters $\Theta$ is given as 

\begin{align}
\Theta^{k+1} = \Theta^{k} - \beta \nabla_{\Theta} \Lc^{pr}(\hat{\wb}^k(\Theta^k)), \label{eq:updateTheta}
\end{align}
where $\beta$ is the learning rate for $\Theta$. This update allows softly selecting useful auxiliary tasks (meta-paths) and balance them with the primary task to improve the performance of the primary task. Without balancing tasks with the weighting function $\Vc(\cdot ; \Theta)$, auxiliary tasks can dominate training and degrade the performance of the primary task.

The model parameters $\wb^k$ for tasks can be updated with optimized $\Theta^{k+1}$ in \eqref{eq:updateTheta} as
\begin{align}
\wb^{k+1} = \wb^{k} - \alpha \nabla_{\wb}\Lc^{pr+au} (\wb^k;\Theta^{k+1}). \label{eq:updateW}
\end{align}
\noindent\textbf{\it{Remarks.}} The proposed formulation can suffer from the meta-overfitting \cite{antoniou2018train,zintgraf2018fast} meaning that the parameters $\Theta$ to learn weights for softly selecting meta-paths and balancing the tasks with the primary task can overfit to the small meta-dataset. 
In our experiment, we found that the overfitting can be alleviated by meta-validation sets \cite{antoniou2018train}. 
To learn $\Theta$ that is generalizable across meta-training sets, we optimize $\Theta$ across $k$ different meta-datasets like $k$-fold  cross validation using the following equation:
\begin{align}
\Theta^{k+1} \myrepeat{1}{\;} = \myrepeat{1}{\;} \underset{D^{pr(meta)}\sim CV\myrepeat{
20}{\;}}{\Theta^{k} \myrepeat{1}{\;} - \myrepeat{2}{\;} \beta \; \;\EE \left [ \; \nabla_{\Theta} \Lc^{pr}(\hat{\wb}^k(\Theta^k))\; \right ],}\label{eq:updateThetaCV}
\end{align}
where $D^{pr(meta)}\sim CV$ is a meta-dataset from cross validation. We used 3-fold cross validation and the gradients of $\Theta$ w.r.t different meta-datasets are averaged to update $\Theta^k$, see Algorithm \ref{alg:selar}. The cross validation is crucial to alleviate meta-overfitting and more discussion is Section \ref{sec:exp_analysis}.
\label{sec:selar}
\subsection{Hint Networks}
\label{sec:hintnet}
\begin{figure}[!htbp]
  \includegraphics[width=0.47\textwidth]{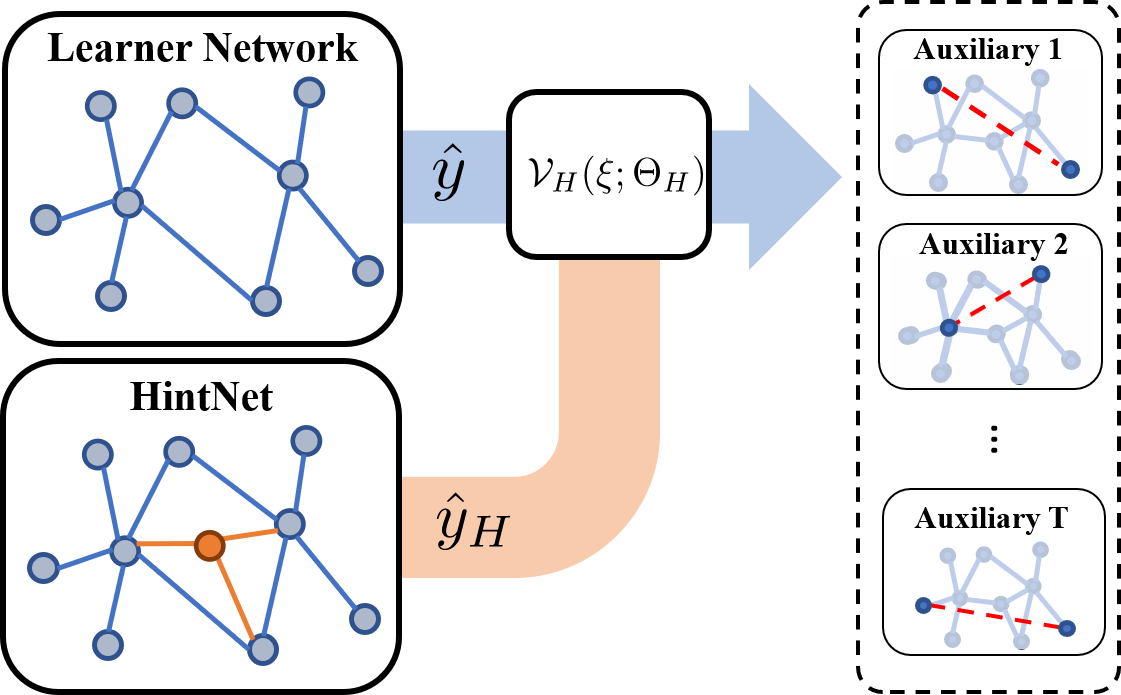}
\caption{HintNet helps the learner network to learn even with challenging and remotely relevant auxiliary tasks. As our framework selects effective auxiliary tasks, our framework with HintNet learns $\Vc_H$ to decide to use hint $\hat{y}_H$ in the \textcolor{orange}{orange} line from HintNet or not via meta-learning. $\hat{y}$ in the \textcolor{blue}{blue} line denotes the prediction from the learner network.}
\label{fig:HintNet}
\end{figure}

Meta-path prediction is generally more challenging than link prediction and node classification since it requires the understanding of long-range relations across heterogeneous nodes.
The meta-path prediction gets more difficult when mini-batch training is inevitable due to the size of datasets or models.
Within a mini-batch, important nodes and edges for meta-paths are not available.
Also, a small learner network, e.g., 2-layer GNNs, with a limited receptive field, inherently cannot capture long-range relations.
The challenges can hinder representation learning and damage the generalization of the primary task.
We proposed a Hint Network (HintNet) which makes the challenging tasks more solvable by correcting the answer with more information at the learner's need. 
Specifically, the HintNet corrects the answer of the learner with its own answer from the augmented graph with hub nodes, see Fig.~\ref{fig:HintNet}. 

The amount of help (correction) by HintNet is optimized maximizing the learner's gain.
Let $\Vc_H(\cdot)$ and $\Theta_H$ be a weight function to determine the amount of hint and its parameters which are optimized by meta-learning. Then, our formulation with HintNet is given as 
\begin{align}
\min_{\wb, \Theta} & \sum_{i=1}^{M_0} \frac{1}{M_0}  \ell^0(y_i^{(0,meta)}, f(x_i^{(0,meta)};\wbst(\Theta, \Theta_H))) \\ 
\text{s.t. } & \wbst(\Theta) = \argmin_{\wb} \sum_{t = 0}^T \sum_{i=1}^{N_t} \frac{1}{N_t} \Vc(\cdot, \ell^t; \Theta) \ell^t(y_i, \hat{y}_i(\Theta_{H})),
\end{align}
where $\hat{y}_i(\Theta_{H})$ denotes the convex combination of the learner's answer and HintNet's answer, i.e., 
$\hat{y}_i(\Theta_{H}) = \Vc_H f^t(x_i^{(t,train)};\wb) + (1-\Vc_H)f_H^t(x_i^{(t,train)};\wb)$ and $\Vc_H = \Vc_H(\xi^{(t, train)}_i; \Theta_H)$.
The input of the weighting function for HintNet is
$\xi^{(t, train)}_i = \left [\ell^t; \ell^t_H; e_t; y_i^{(t,train)} \right ] \in \Rb^{T+3}$.

\section{Experiments}
We evaluate our proposed methods on four public benchmark datasets of heterogeneous graphs.
Our experiments answer the following research questions: \textbf{Q1.} Is meta-path prediction effective for representation learning on heterogeneous graphs?
\textbf{Q2.} Can the meta-path prediction be further improved by the proposed methods (e.g., SELAR, HintNet)?
\textbf{Q3.} Can our proposed methods benefit from various auxiliary tasks in the primary task performance?
\textbf{Q4.} Why are the proposed methods effective, any relation with hard negative mining? 

\subsection{Settings}

\hspace{\parindent} \textbf{Datasets.} We use two public benchmark datasets from different domains for link prediction: Music dataset Last-FM and Book dataset Book-Crossing, released by KGNN-LS~\cite{wang2019knowledge}, RippleNet~\cite{wang2018ripplenet}.
The Last-FM dataset with a knowledge graph has 122 types of edges, e.g., "artist.origin", "musician.instruments.played", "person.or.entity.appearing.in.film", and "film.actor.film", etc.
Book-Crossing with a knowledge graph has 52 types of edges, e.g., "book.genre", "literary.series", "date.of.first.publication", and "written.work.translation", etc.

%%%%%%%%%%%%%%%%%%%%%%%%%%%%%%%%%%%%%%%
\begin{table}[ht]
\centering
\caption{Datasets on heterogeneous graphs.}
% \footnotesize 
\label{tab:data}
\resizebox{\columnwidth}{!}{\begin{tabular}{lrrrr}
\toprule
  Datasets & \# Nodes & \# Edges & \# Edge type & \# Features \\
 \midrule
 Last-FM    & 15,084  & 73,382  & 122 & N/A\\
 Book-Crossing  & 110,739 & 442,746 & 52 & N/A  \\
\midrule
 ACM    & 8,994  & 25,922 & 4 & 1,902   \\
 IMDB  & 12,772  & 37,288 & 4 & 1,256  \\
\bottomrule
\end{tabular}}
\end{table}

% \begin{table*}[h]
% \centering
% \caption{Datasets on heterogeneous graphs.}
% \label{tab:node}
% \begin{tabular}{ccccccc}
% \hline
% \textbf{Dataset} & \ \textbf{\# Nodes} & \ \textbf{\# Edges} & \ \textbf{\#Edge Type} & \ \textbf{\# Features}  \\ \hline
% Last-FM    & 15,084  & 42,346  & 60 & N/A\\
% \\[-0.8em]
% Book-Crossing  & 110,739 & 139,746 & 25 & N/A  \\
% \\[-0.8em]
% ACM    & 8,994  & 25,922 & 4 & 1,902   \\
% \\[-0.8em]
% IMDB  & 12,772  & 37,288 & 4 & 1,256  \\
% \end{tabular}
% \end{table*}

% \begin{table*}[h]
% \centering
% \caption{Datasets for link prediction on heterogeneous graphs.}
% \label{tab:edge}
% \begin{tabular}{ccccccc}
% \hline
% \textbf{Dataset} & \ \textbf{\# Users} & \ \textbf{\# Items} & \ \textbf{\# Interactions} & \ \textbf{\# Entity} & \ \textbf{\# Relations} \\ \hline
% Last-FM    & 1,872  & 3,846 & 42,346 & 9,366 & 60 \\
% \\[-0.8em]
% Book-Crossing  & 17,860  & 14,976 & 139,746 & 77,903 & 25 \\
% \end{tabular}
% \end{table*}

% \begin{table*}[h]
% \centering
% \caption{Datasets for node classification on heterogeneous graphs.}
% \label{tab:node}
% \begin{tabular}{ccccccc}
% \hline
% \textbf{Dataset} & \ \textbf{\# Nodes} & \ \textbf{\# Edges} & \ \textbf{\#Edge Type} & \ \textbf{\# Features} & \ \textbf{ Meta-path} \\ \hline
% ACM    & 8,994  & 25,922 & 4 & 1,902  & PAP, PSP  \\
% \\[-0.8em]
% IMDB  & 12,772  & 37,288 & 4 & 1,256  & MAM, MDM\\
% \end{tabular}
% \end{table*}
\begin{table*}[ht]
  \centering
  \caption{Link prediction performance ($AUC$) of GNNs trained by various learning strategies.}
  \label{tab:link prediction}
  \begin{tabular}{c@{ }c|rr|rrr}
    \toprule
    \multirow{2}*{Dataset}&\multirow{2}*{Base GNNs}&\multirow{2}*{Vanilla}&\multirow{2}{*}{\shortstack{w/o \\  meta-path}} & \multicolumn{3}{c}{Ours} \\
    %\hline
    & & & & w/ meta-path & SELAR & SELAR+Hint \\
    %\toprule
    \midrule
    \multirow{6}{*}{Last-FM} 
    & GCN & 0.7963  & 0.7889 & 0.8235 & \textbf{0.8296} & 0.8121  \\
    & GAT & 0.8115  & 0.8115 & 0.8263 & 0.8294 & \textbf{0.8302}\\
    & GIN & 0.8199  & 0.8217 & 0.8242 & \textbf{0.8361} & 0.8350\\
    & SGC & 0.7703  & 0.7766 & 0.7718 & 0.7827 & \textbf{0.7975} \\
    & GTN & 0.7836  & 0.7744 & 0.7865 & 0.7988 & \textbf{0.8067} \\
    \cmidrule{2-7}
    & Avg. Gain & - & -0.0017 & +0.0106 & +0.0190 & +0.0200 \\
    \midrule
    \multirow{6}{*}{Book-Crossing} 
    & GCN & 0.7039 & 0.7031 & 0.7110 & 0.7182 & \textbf{0.7208}\\
    & GAT & 0.6891 & 0.6968  & 0.7075 & 0.7345 & \textbf{0.7360} \\
    & GIN & 0.6979 & 0.7210  & 0.7338 & \textbf{0.7526} & 0.7513\\
    & SGC & 0.6860 & 0.6808 & 0.6792 & 0.6902 & \textbf{0.6926} \\
    & GTN & 0.6732 & 0.6758 & 0.6724 & \textbf{0.6858} & 0.6850 \\
    \cmidrule{2-7}
    & Avg. Gain & - & +0.0055 & +0.0108 & +0.0263 & +0.0267\\
    \bottomrule
  \end{tabular}
\end{table*}
%\begin{table*}[h]
%\centering
%\caption{Task-specific weights on \textbf{Last-FM} and \textbf{Book-Crossing} datasets}
%\label{tab:node}
%\begin{tabular}{|l|c||l|c|}
%\hline
%\hline
%\textbf{meta-path} & \textbf{Avg. weighted loss} \\
%$user \xrightarrow{t_1} item \xrightarrow{t_2} actor \xrightarrow{item} item$ &  \\
%\hline
%\end{tabular}
%\end{table*}

\setlength{\tabcolsep}{10pt}
\begin{table*}[ht]
\centering
\caption{The average of the task-specific weighted loss.}
\label{tab:MP}
\begin{tabular}{lc|lc}
\toprule
Meta-paths (Last-FM) & Avg. & Meta-paths (Book-Crossing) & Avg. \\
\midrule
user-item-actor-item & \textbf{7.675} & user-item$^{*}$ & \textbf{6.439} \\
user-item$^{*}$ & 7.608 & user-item-literary.series-item-user & 6.217\\
user-item-appearing.in.film-item & 7.372 & item-genre-item & 6.163\\
user-item-instruments-item & 7.049 & user-item-user-item & 6.126\\
user-item-user-item & 6.878 & user-item-user & 6.066\\
user-item-artist.origin-item & 6.727 & item-user-item & 6.025\\
\bottomrule
{\raggedright ${*}$ primary task \par}
\end{tabular}
\end{table*}
%%%%%%%%%%%%%%%%%%%%%%%%%%%%%%%%%%%%%%%

We use two datasets for node classification: citation network datasets ACM and Movie dataset IMDB, used by HAN~\cite{HAN} for node classification tasks.
ACM has three types nodes (Paper(P), Author(A), Subject(S)), four types of edges (PA, AP, PS, SP) and labels (categories of papers). 
IMDB contains three types of nodes (Movie (M), Actor (A), Director (D)),  four types (MA, AM, MD, DM) of edges  and labels (genres of movies). 
Last-FM and Book-Crossing do not have node features, while ACM and IMDB have node features, which are bag-of-words of keywords and plots.
Statistics of the datasets are in Table~\ref{tab:data}.

{\textbf{Baselines.} }
We evaluate our methods with five graph neural networks : GCN~\cite{GCN}, GAT~\cite{GAT}, GIN~\cite{xu2018powerful}, SGConv~\cite{wu2019simplifying} and GTN~\cite{yun2019graph}. Our methods can be applied to both homogeneous graphs and heterogeneous graphs. We compare five learning strategies: \textbf{Vanilla}, standard training of base models only with the primary task samples; \textbf{w/o meta-path}, learning a primary task with sample weighting function $\Vc(\xi;\Theta)$; \textbf{w/ meta-path}, training with the primary task and auxiliary tasks (meta-path prediction) with a standard loss function; \textbf{SELAR} proposed in Section \ref{sec:selar}, learning the primary task with optimized auxiliary tasks by meta-learning; \textbf{SELAR+Hint} introduced in Section \ref{sec:hintnet}.

{\textbf{Implementation details.} }
All the models are randomly initialized and optimized using Adam~\cite{DBLP:journals/corr/KingmaB14} optimizers.
For a fair comparison, the number of layers is set to two and the dimensionality of output node embeddings is the same across models.
The node embedding $z$ for Last-FM has 16 dimensions and for the rest of the datasets 64 dimensions.
Since datasets have a different number of samples, we train models for a different number of epochs; Last-FM (100), Book-Crossing (50), ACM (200), and IMDB (200). 
For link prediction, the neighborhood sampling algorithm~\cite{graphsage2017} is used and the neighborhood size is 8 and 16 in Last-FM and Book-Crossing respectively.
For node classification, the neighborhood size is 8 in all datasets.
For the experiments on link prediction using \textbf{SELAR+Hint}, we train a learner network with attenuated weights as a regularization, i.e., $\Vc_H(\xi^{(t, train)}_i; \Theta_H)^\gamma f^t(x_i^{(t,train)};\wb) + (1-\Vc_H(\xi^{(t, train)}_i; \Theta_H)^\gamma)f_H^t(x_i^{(t,train)};\wb)$ , where $0 < \gamma \le 1$.
In all the experiments, hyperparameters such as learning rate and weight-decay rate are tuned using validation sets for all models. 
The test performance was reported with the best models on the validation sets.
We report the mean performance of three independent runs. 
Our experiments were mainly performed based on NAVER Smart Machine Learning platform (NSML)~\cite{sung2017nsml,kim2018nsml}. %\vspace{0.15cm}
All the experiments use PyTorch~\cite{paszke2017automatic} and the geometric deep learning extension library provided by Fey \& Lenssen~\cite{Fey:2019wv}.
%%%%%%%%%%%%%%%%%%%%%%%%%%%%%%%%%%%%%%%%%%%%%
% \begin{table*}[ht]
%   \centering
%   \caption{The results of \textit{Recall@K} in top-K recommendation.}
%   \label{tab:recommender}
%   \begin{tabular}{l|cccc}
%     \toprule
%     Model  & \textit{Recall@2} & \textit{Recall@10} & \textit{Recall@50} & \textit{Recall@100}  \\
%     \midrule
%     SVD  & 0.029 & 0.098 & 0.240 & 0.332    \\
%     LibFM & 0.030 & 0.103 & 0.263 & 0.330   \\
%     LibGM + TransE  & 0.032 & 0.102 & 0.259 & 0.326 \\ 
%     PER & 0.014 & 0.052 & 0.116 & 0.176  \\
%     CKE  & 0.023 & 0.070 & 0.180 & 0.296  \\
%     RippleNet & 0.032 & 0.101 & 0.242 & 0.336 \\
%     KGNN-LS & 0.044 & \textbf{0.122} & 0.277 & 0.370  \\
%     \cmidrule{1-5}
%     Vanilla (GCN) & 0.018 & 0.070 & 0.174 & 0.223   \\
%     Vanilla (GAT)  & 0.010 & 0.071 & 0.200 & 0.311  \\
%     Vanilla (GIN) & 0.018 & 0.059 & 0.266 & 0.330 \\
%     Vanilla (SGC) & 0.005 & 0.030 & 0.203 & 0.321  \\
%     Vanilla (GTN) & 0.012 & 0.071 & 0.175 & 0.264 \\
%     \cmidrule{1-5}
%      SELAR (GCN) & 0.028 & 0.084 & 0.238 & 0.317  \\
%      SELAR (GAT) & \textbf{0.046} & 0.119 & 0.297 & 0.402 \\
%      SELAR (GIN) & 0.025 & 0.033 & 0.275 & 0.371 \\
%      SELAR (SGC) & 0.003 & 0.051 & 0.251 & 0.353   \\
%      SELAR (GTN) & 0.002 & 0.097 & \textbf{0.345} & \textbf{0.415}  \\
%     \bottomrule
%   \end{tabular}
% \end{table*} 

\begin{table*}[ht]
  \centering
  \caption{Comparison with various baselines in top-K recommendation (\textit{Recall@K}).}
  %The results of \textit{Recall@K} with various baselines in top-K recommendation}
  %Comparison (\textit{Recall@K}) with recommendation baselines and GNNs baselines in top-K recommendation.}
  \label{tab:recommender}
  \begin{tabular}{cl|cccc}
    \toprule
    Dataset & Model  & \textit{Recall@2} & \textit{Recall@10} & \textit{Recall@50} & \textit{Recall@100}  \\
    \midrule
    \multirow{17}{*}{Last-FM} 
    & SVD  & 0.029 & 0.098 & 0.240 & 0.332    \\
    & LibFM & 0.030 & 0.103 & 0.263 & 0.330   \\
    & LibGM + TransE  & 0.032 & 0.102 & 0.259 & 0.326 \\\
    & PER  & 0.014 & 0.052 & 0.116 & 0.176  \\
    & CKE  & 0.023 & 0.070 & 0.180 & 0.296  \\
    & RippleNet & 0.032 & 0.101 & 0.242 & 0.336 \\
    & KGNN-LS & 0.044 & \textbf{0.122} & 0.277 & 0.370  \\
    \cmidrule{2-6}
    & Vanilla (GCN) & 0.010 & 0.018 & 0.238 & 0.291   \\
    & Vanilla (GAT)  & 0.027 & 0.082 & 0.269 & 0.320  \\
    & Vanilla (GIN) & 0.018 & 0.059 & 0.266 & 0.330 \\
    & Vanilla (SGC) & 0.001 & 0.044 & 0.193 & 0.264  \\
    & Vanilla (GTN) & 0.012 & 0.071 & 0.175 & 0.264 \\
    \cmidrule{2-6}
    & SELAR (GCN) & 0.014 & 0.059 & 0.298 & \textbf{0.416}  \\
    & SELAR (GAT) & \textbf{0.046} & 0.119 & 0.297 & 0.402 \\
    & SELAR (GIN) & 0.015 & 0.063 & 0.303 & 0.390 \\
    & SELAR (SGC) & 0.003 & 0.051 & 0.251 & 0.353   \\
    & SELAR (GTN) & 0.002 & 0.097 & \textbf{0.345} & 0.415  \\

    \bottomrule
  \end{tabular}
\end{table*}

% \begin{table*}[ht]
%   \centering
%   \caption{The results of \textit{Recall@K} in top-K recommendation.}
%   \begin{tabular}{cl|rrrr}
%     \toprule
%     Dataset & Base models  & \textit{R@2} & \textit{R@10} & \textit{R@50} & \textit{R@100}  \\
%     \midrule
%     \multirow{12}{*}{Last-FM} 
%     & SVD  & 0.029 & 0.098 & 0.240 & 0.332    \\
%     & LibFM & 0.030 & 0.103 & 0.263 & 0.330   \\
%     & LibGM + TransE  & 0.032 & 0.102 & 0.259 & 0.326 \\\
%     & PER  & 0.014 & 0.052 & 0.116 & 0.176  \\
%     & CKE  & 0.023 & 0.070 & 0.180 & 0.296  \\
%     & RippleNet & 0.032 & 0.101 & 0.242 & 0.336 \\
%     & KGNN-LS & 0.044 & 0.122 & 0.277 & 0.370  \\

%     \cmidrule{2-6}
%     & GCN & 0.018 & 0.070 & 0.174 & 0.223   \\
%     & SELAR (GCN) & 0.028 & 0.084 & 0.238 & 0.317  \\
%     & GAT  & 0.010 & 0.071 & 0.200 & 0.311  \\
%     & SELAR (GAT) & \textbf{0.046} & 0.119 & \textbf{0.297} & \textbf{0.402} \\
%     & GIN & 0.018 & 0.059 & 0.266 & 0.330 \\
%     & SELAR (GIN) & 0.025 & 0.033 & 0.275 & \textbf{0.371} \\
%     & SGC & 0.005 & 0.030 & 0.203 & 0.321  \\
%     & SELAR (SGC) & 0.003 & 0.051 & 0.251 & 0.353   \\
%     & GTN & 0.012 & 0.071 & 0.175 & 0.264 \\
%     & SELAR (GTN) & 0.002 & 0.097 & \textbf{0.345} & \textbf{0.415}  \\

%     \bottomrule
%   \end{tabular}
% \end{table*} 
    % & Vanilla (GCN) & 0.018 & 0.070 & 0.174 & 0.223   \\
    % & Vanilla (GAT)  & 0.010 & 0.071 & 0.200 & 0.311  \\
    % & Vanilla (GIN) & 0.018 & 0.059 & 0.266 & 0.330 \\
    % & SELAR (GCN) & 0.028 & 0.084 & 0.238 & 0.317  \\
    % & SELAR (GIN) & 0.025 & 0.033 & 0.275 & 0.371 \\

\begin{table*}[ht]
  \centering
  \caption{Node classification performance ($F1$-score) of GNNs trained by various learning schemes.}
  \label{tab:node classification}
  \begin{tabular}{cc|rr|rrr}
    \toprule
    \multirow{2}*{Dataset}&\multirow{2}*{Base GNNs}&\multirow{2}*{Vanilla}&\multirow{2}{*}{\shortstack{w/o \\  meta-path}} & \multicolumn{3}{c}{Ours} \\
    & & & & w/ meta-path & SELAR & SELAR+Hint \\
    \midrule
    \multirow{6}{*}{ACM} & GCN & 0.9091 & 0.9191 & 0.9104 & 0.9229 & \textbf{0.9246} \\
    & GAT & 0.9161 & 0.9119 & 0.9262 & 0.9273 & \textbf{0.9278} \\
    & GIN & 0.9085 & 0.9118 & 0.9058 & 0.9092 & \textbf{0.9135}\\
    & SGC & 0.9163 & 0.9194 & 0.9223 & 0.9224 & \textbf{0.9235} \\
    & GTN & 0.9181 & 0.9191 & 0.9246 & \textbf{0.9258}  &  0.9236\\
    \cmidrule{2-7}
    & Avg. Gain & - & +0.0027 & +0.0043 & +0.0079 & \textbf{+0.0090} \\
    \midrule
    \multirow{6}{*}{IMDB} & GCN & 0.5767 & 0.5855 & 0.5994 & 0.6083 & \textbf{0.6154}\\
    & GAT & 0.5653 & 0.5488 & 0.5910 & \textbf{0.6099} & 0.6044 \\
    & GIN & 0.5888 & 0.5698 & 0.5891 & \textbf{0.5931} & 0.5897 \\
    & SGC & 0.5779 & 0.5924 & 0.5940 & 0.6151 & \textbf{0.6192} \\
    & GTN & 0.5804 & 0.5792 & 0.5818 & 0.5994 & \textbf{0.6063} \\
    \cmidrule{2-7}
    & Avg. Gain & - & -0.0027 & +0.0132 & +0.0274 & \textbf{+0.0292} \\
    \bottomrule
  \end{tabular}
\end{table*} 

%%%%%%%%%%%%%%%%%%%%%%%%%%%%%%%%%%%%%%%%%%%%%
\subsection{Link Prediction with meta-path prediction}
{\;\;\;\; \textbf{Link prediction results.} }
We used five types of meta-paths of length 2 to 4 for auxiliary tasks.
Table~\ref{tab:link prediction} shows that our methods consistently improve link prediction performance for all the GNNs, compared to the Vanilla and the method using Meta-Weight-Net~\cite{han2018coteaching} only without meta-paths (denoted as w/o meta-path).
Overall, a standard training with meta-paths shows 1.1\% improvement on average on both Last-FM and Book-Crossing whereas meta-learning that learns sample weights degrades on average on Last-FM and improves only 0.6\% on average on Book-Crossing, e.g., GCN, SGC and GTN on Last-FM and GCN and SGC on Book-Crossing, show degradation 0.2\% compared to the standard training (Vanilla). 
As we expected, SELAR and SELAR with HintNet provide more optimized auxiliary learning resulting in  1.9\% and 2.0\% absolute improvement on Last-FM and 2.6\% and 2.7\% on the Book-Crossing dataset.
Further, in particular, GIN on Book-crossing, SELAR and SELAR+Hint provide $\sim$5.5\% and $\sim$5.3\% absolute improvement compared to the vanilla algorithm.

{\textbf{Task selection results.} }
Our proposed methods can identify useful auxiliary tasks and balance them with the primary task. 
In other words, the loss functions for tasks are differentially adjusted by the weighting function learned via meta-learning.
To analyze the weights of the tasks, we calculate the average of the task-specific weighted loss learned by SELAR+HintNet for GAT.
Table. \ref{tab:MP} shows tasks in descending order of the task weights. `user-item-actor-item' has the largest weight followed by `user-item' (primary task), `user-item-appearing.in.film-item', `user-item-instruments-item', `user-item-user-item' and `user-item-artist.origin-item' on the Last-FM. 
It indicates that the preference of a given user is closely related to other items connected by an actor, e.g., specific edge type `film.actor.film' in the knowledge graph. 
Moreover, our method focuses on `user-item' interaction for the primary task.
On the Book-Crossing dataset, our method has more attention to `user-item' for the primary task and `user-item-literary.series-item-user' which means that users who like a series book have similar preferences. 
%%%%%%%%%%%%%%%%%%%%%%%%%%%%%%%%%%%%%%%%%%%%%
%%%%%%%%%%%%%%%%%%%%%%%%%%%%%%%%%%%%%%%%%%%%%
\subsection{Recommender system with meta-path prediction}
To evaluate our framework in the recommendation scenario on the music dataset Last-FM, we use top-K recommendation Recall@K, which is the proportion of the relevant items among those selected as the $K$ items with the highest predicted probability for each user. 
We compare our proposed method with the baseline methods to demonstrate its effectiveness in the recommendation task. 
We use SVD~\cite{koren2008factorization}, LibFM~\cite{rendle2012factorization}, LibFM+TransE~\cite{rendle2012factorization, bordes2013translating}, PER~\cite{yu2014personalized}, CKE~\cite{zhang2016collaborative}, RippleNet~\cite{wang2018ripplenet}, and KGNN-LS~\cite{wang2019knowledge} as recommendation baselines and GCN~\cite{GCN}, GAT~\cite{GAT}, GIN~\cite{xu2018powerful}, SGC~\cite{wu2019simplifying}, and GTN~\cite{yun2019graph} as GNN base models. 
For recommendation baselines, we directly use the results reported in~\cite{wang2019knowledge}.
The experimental results in top-K recommendation are presented in Table.~\ref{tab:recommender}.
SELAR consistently improves the performance on Last-FM, compared to the vanilla learning scheme. 
In particular, SELAR improved the Recall@2 of GAT by 1.9\% absolute gain compared to the vanilla. 
SELAR on GTN and GCN shows 17\% and 13\% improvements from the vanilla in Recall@50 and Recall@100, respectively.
Also, SELAR performed as much as KGNN-LS, a model designed for user-specific recommendations, and yielded the best performance in the rest of the evaluation metrics except Recall@10.

\begin{table*}[!ht]
  \centering
%   \footnotesize
  \caption{Node classification performance ($F1$-score) of GNNs trained by various combinations of the auxiliary tasks.}
  \small
  \label{tab:multiple}
\begin{threeparttable}
 \begin{tabular}{c|l|l|l}
    \toprule
    Base GNNs & Combination of auxiliary tasks &  \multicolumn{1}{c}{MTL}  &   \multicolumn{1}{c}{SELAR} \\
    \midrule
    \multirow{12}{*}{GCN}
    & - & 0.5767 & \\
    & Meta-path & 0.5994 & \textbf{0.6083}\\
    & Degree & 0.5886 & \textbf{0.6074} \\
    & Distance & 0.5880 & \textbf{0.6080}\\
    & PageRank & 0.5906  & \textbf{0.5919} \\
    & Partition & 0.5835 & \textbf{0.6127} \\
    & Clustering & 0.5841 & \textbf{0.5878} \\
    & Degree+Distance & 0.5975 & \textbf{0.5983}\\
    & Degree+Distance+PageRank & 0.5908 & \textbf{0.5993}\\
    & Degree+Distance+PageRank+Partition & 0.5864 & \textbf{0.6255}\\
    & Degree+Distance+PageRank+Meta-path & 0.6162 & \textbf{0.6317}\\
    & Degree+Distance+PageRank+Partition+Clustering+Meta-path & \textbf{0.6096} & 0.6036\\
    \cmidrule{1-4}
    \multirow{12}{*}{GAT}
    & - & 0.5653 & \\
    & Meta-path & 0.5910 & \textbf{0.6099}\\
    & Degree & 0.5677 & \textbf{0.5798}\\
    & Distance & \textbf{0.5724} & 0.5699\\
    & PageRank & 0.5479$^{*}$ & \textbf{0.5801}\\
    & Partition & 0.5612$^{*}$ & \textbf{0.5672}\\
    & Clustering & 0.5589$^{*}$ & \textbf{0.5663}\\
    & Degree+Distance & 0.5656 & \textbf{0.5836}\\
    & Degree+Distance+PageRank & 0.5741 & \textbf{0.6010}\\
    & Degree+Distance+PageRank+Partition & 0.5910 & \textbf{0.6115}\\
    & Degree+Distance+PageRank+Meta-path  & 0.6029 & \textbf{0.6164}\\
    & Degree+Distance+PageRank+Partition+Clustering+Meta-path & \textbf{0.6132} & 0.6100\\
    \cmidrule{1-4}
    \multirow{12}{*}{GIN}
    & - & 0.5888 & \\
    & Meta-path & 0.5891 & \textbf{0.5931}\\
    & Degree & 0.5814$^{*}$ & \textbf{0.5936}\\
    & Distance & 0.5827$^{*}$ & \textbf{0.5849}$^{*}$\\
    & PageRank & 0.5882$^{*}$ & \textbf{0.5902}\\
    & Partition & 0.5808$^{*}$ & \textbf{0.5859}$^{*}$ \\
    & Clustering & \textbf{0.5816}$^{*}$ & 0.5812$^{*}$ \\
    & Degree+Distance & 0.5800$^{*}$ & \textbf{0.5878}$^{*}$\\
    & Degree+Distance+PageRank & 0.5813$^{*}$ & \textbf{0.5917}\\
    & Degree+Distance+PageRank+Partition & 0.5790$^{*}$& \textbf{0.5841}$^{*}$\\
    & Degree+Distance+PageRank+Meta-path  & 0.5963$^{*}$ & \textbf{0.6039}\\
    & Degree+Distance+PageRank+Partition+Clustering+Meta-path & 0.5754$^{*}$ & \textbf{0.5933}\\
    \cmidrule{1-4}
    \multirow{12}{*}{SGC}
    & - & 0.5779 &  \\
    & Meta-path & 0.5940 & \textbf{0.6151}\\
    & Degree & 0.5798 & \textbf{0.5906}\\
    & Distance & 0.5723$^{*}$ & \textbf{0.5998}\\
    & PageRank & 0.5934 & \textbf{0.5940}\\
    & Partitioning & 0.5982 & \textbf{0.6062}\\
    & Clustering & \textbf{0.5933} & 0.5902\\
    & Degree+Distance & 0.5872 & \textbf{0.5951}\\
    & Degree+Distance+PageRank & 0.5833 & \textbf{0.5906}\\
    & Degree+Distance+PageRank+Partitioning & 0.5871 & \textbf{0.5960}\\
    & Degree+Distance+PageRank+Meta-path & 0.5868 & \textbf{0.6091}\\
    & Degree+Distance+PageRank+Partitioning+Clustering+Meta-path & 0.5897 & \textbf{0.5968}\\
    \cmidrule{1-4}
    \multirow{12}{*}{GTN}
    & - & 0.5804 & \\
    & Meta-path & 0.5818 & \textbf{0.5994}\\
    & Degree & 0.6152 & \textbf{0.6161}\\
    & Distance & 0.6100 & \textbf{0.6387}\\
    & PageRank & 0.6029 & \textbf{0.6184}\\
    & Partition & \textbf{0.6288} & 0.6224\\
    & Clustering & \textbf{0.6170} & \textbf{0.6170}\\
    & Degree+Distance & \textbf{0.6270} & 0.6229 \\
    & Degree+Distance+PageRank & 0.5966 & \textbf{0.6149}\\
    & Degree+Distance+PageRank+Partition & 0.6143 & \textbf{0.6199}\\
    & Degree+Distance+PageRank+Meta-path & 0.6135 & \textbf{0.6218} \\
    & Degree+Distance+PageRank+Partition+Clustering+Meta-path & 0.6152 & \textbf{0.6184}\\
    \bottomrule
    \end{tabular}
    \begin{tablenotes}
    \footnotesize
    \item \hspace{-0.2cm} ${*}$ indicates the performance degradation compared to Vanilla.
    \end{tablenotes}
    \end{threeparttable}
\end{table*}

    % & - & \multicolumn{2}{c}{0.5888}\\

 \subsection{Node Classification with meta-path prediction}
Similar to link prediction above, Table~\ref{tab:node classification} shows that our SELAR consistently enhances node classification performance of all the GNN models and the improvements are more significant on IMDB which is larger than the ACM dataset. 
We believe that ACM dataset is already saturated and the room for improvement is limited. However, our methods still show small yet consistent improvement over all the architecture on ACM. 
We conjecture that the efficacy of our proposed methods differs depending on graph structures. 
However, it is worth noting that introducing meta-path prediction as auxiliary tasks remarkably improves the performance of primary tasks such as link and node prediction with consistency compared to the existing methods. ``w/o meta-path'', the meta-learning to learn sample weight function on a primary task shows marginal degradation in five out of eight settings.
Remarkably, SELAR improved the F1-score of GAT on the IMDB by (4.46\%) compared to the vanilla learning scheme.

%%%%%%%%%%%%%%%%%%%%%%%%%%%%%%%%%%%%%%%%%%%%%
\begin{figure*}[!htbp]
\centering
\subfigure[Weighting function $\Vc(\xi;\Theta)$.]{
	\includegraphics[height=4.7cm]{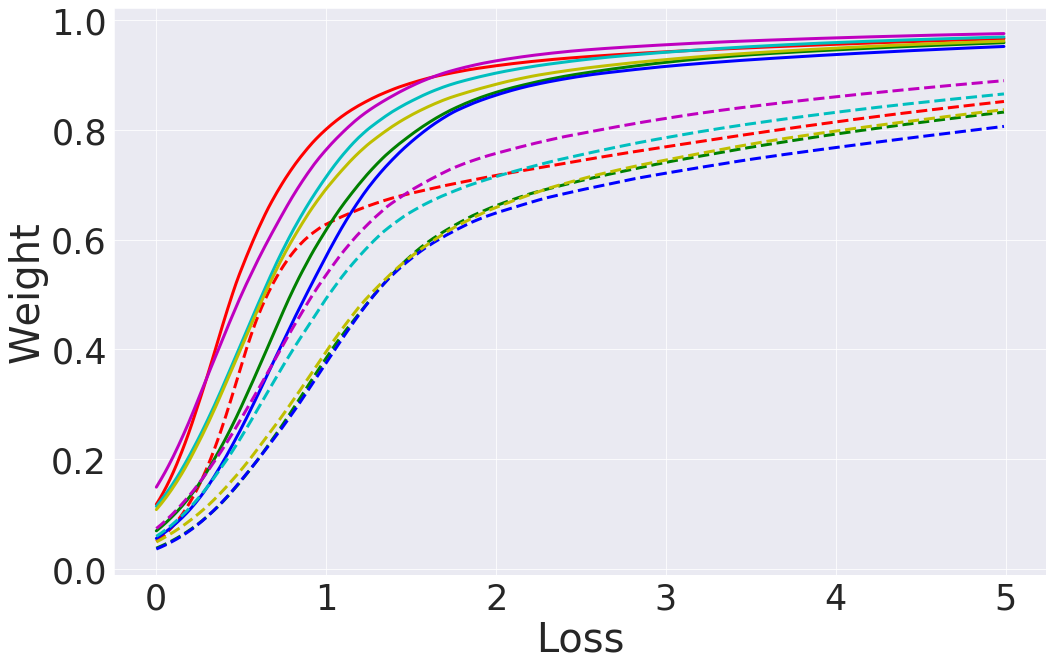}\label{fig:weighfunc_a}
    }
\subfigure[Adjusted Cross Entropy $\Vc(\xi;\Theta)\ell^t(y,\hat{y})$.]{
\includegraphics[height=4.7cm]{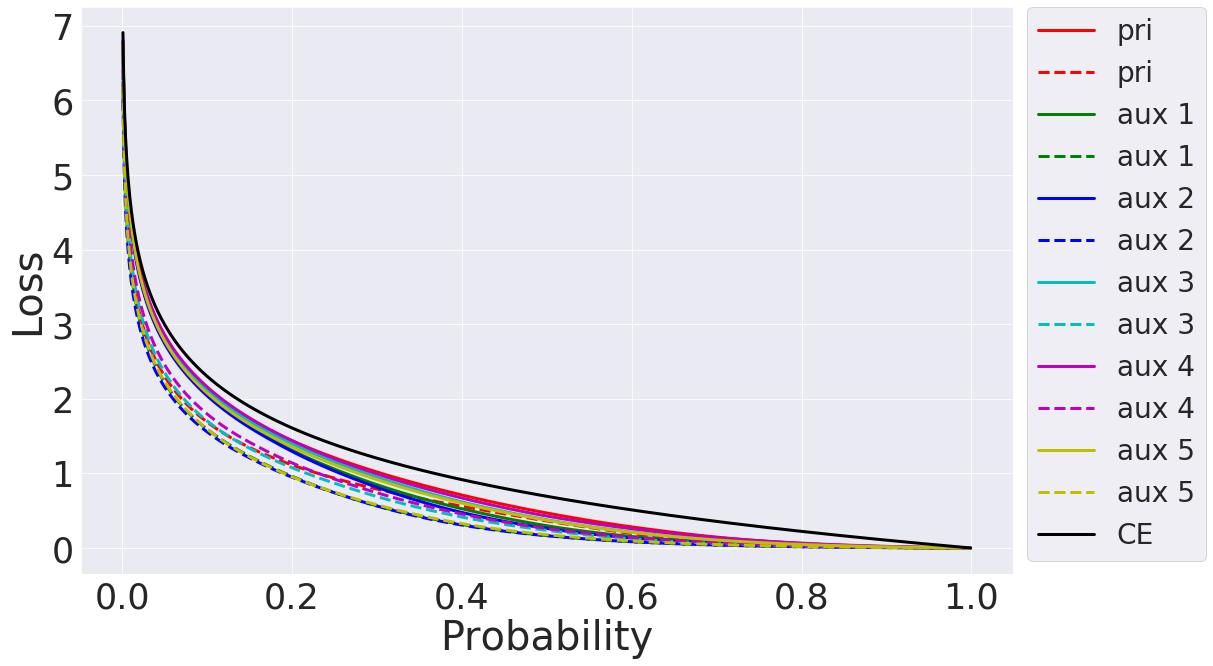}
}
\caption{Weighting function $\Vc(\cdot)$ learnt by SELAR+HintNet. $\Vc(\cdot)$ gives overall high weights to the primary task positive samples (\textcolor{red}{red}) in (a). $\Vc(\cdot)$ decreases the weights of easy samples with a loss ranged from 0 to 1. In (b), the adjusted cross entropy, i.e., $-\Vc(\xi;\Theta)\log(\hat{y})$, by $\Vc(\cdot)$ acts like the focal loss, which focuses on hard examples by $-(1-p_t)^{\gamma}\log(\hat{y})$.
% $\Vc(\cdot)$ adjusts the cross entropy (CE), \textcolor{black}{black} line in (b), as $-\Vc(\xi;\Theta)\log(\hat{y})$, which is similar to the focal loss. The adjusted loss focuses on hard samples.
}
\label{fig:weighfunc}
\end{figure*}

\begin{figure*}[!htbp]
\subfigure[GCN]{
    	\includegraphics[trim=0 0 210 0, clip,height=3.8cm]{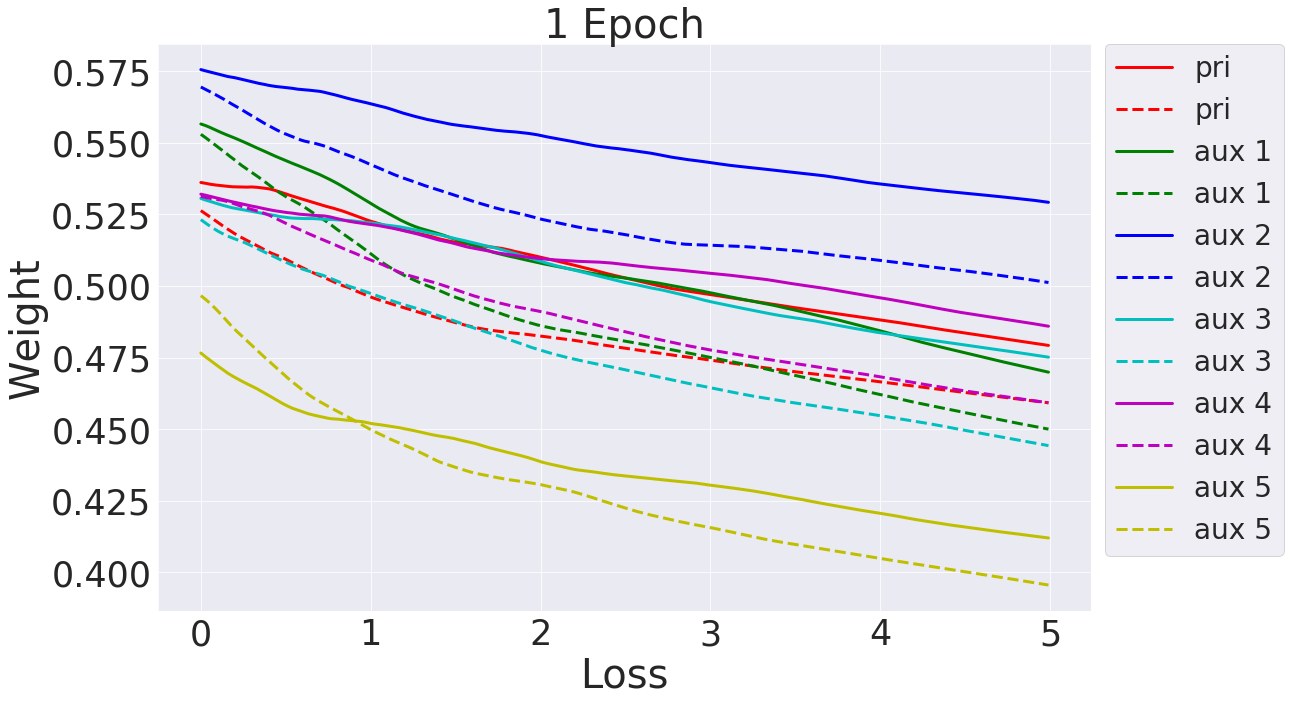}%
    	\includegraphics[trim=0 0 210 0, clip,height=3.8cm]{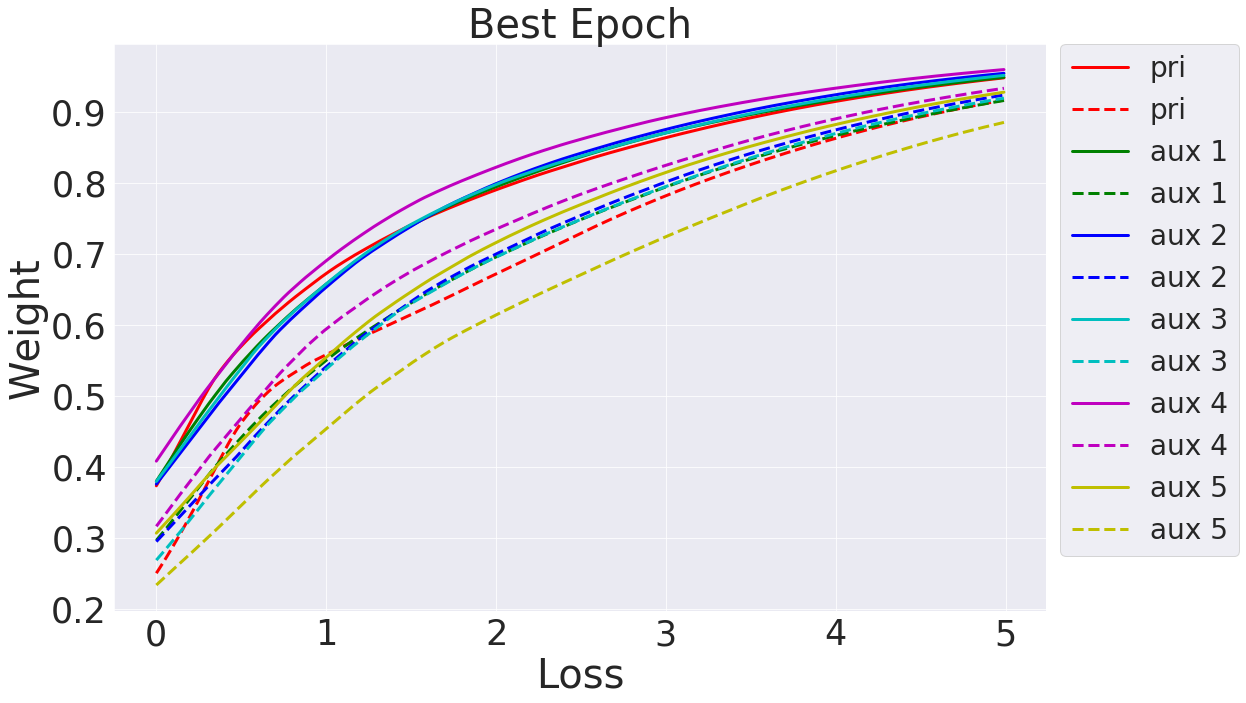}%
    	\includegraphics[height=3.8cm]{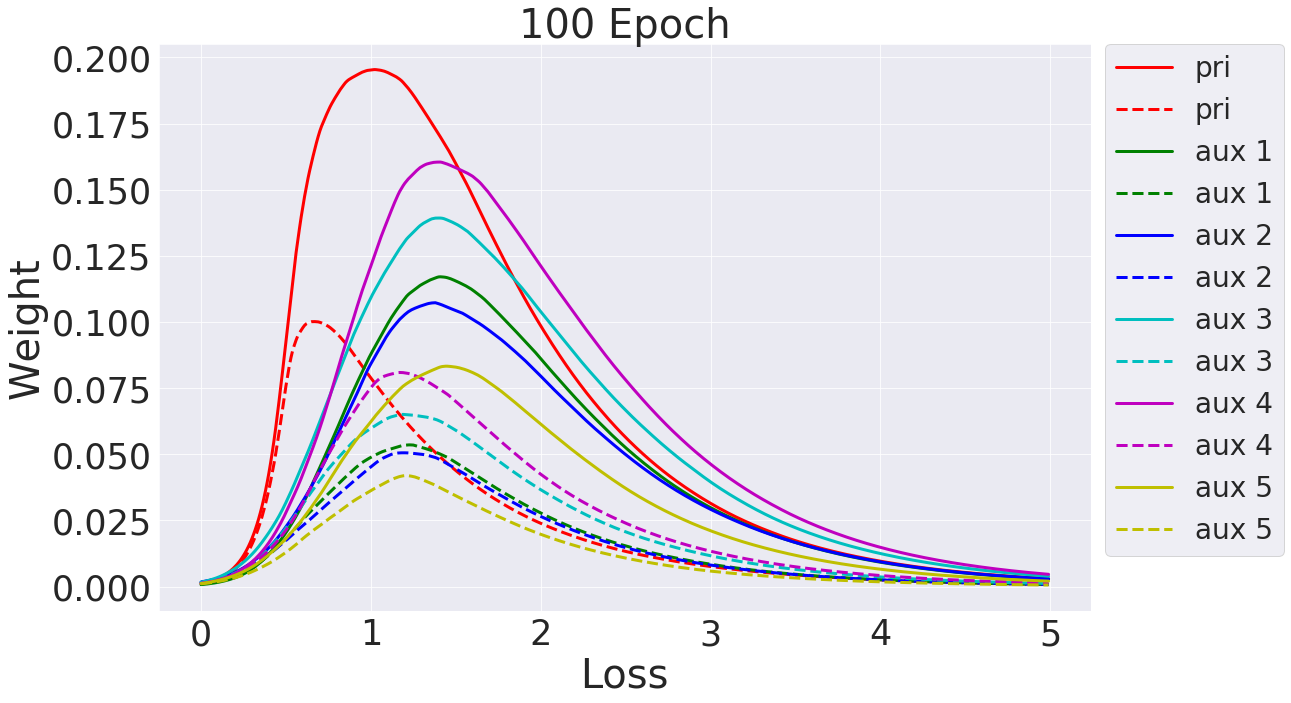}
    	}
\subfigure[GAT]{
    	\includegraphics[trim=0 0 210 0, clip,height=3.7cm]{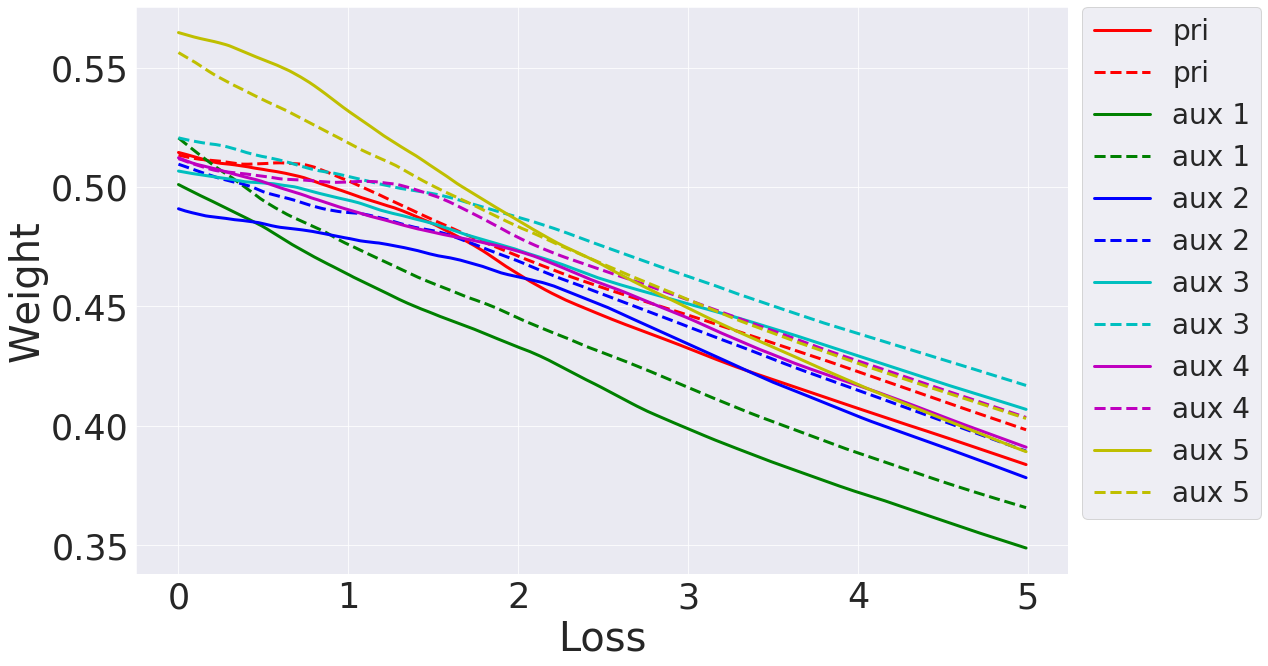}%
    	\includegraphics[trim=0 0 210 0, clip,height=3.7cm]{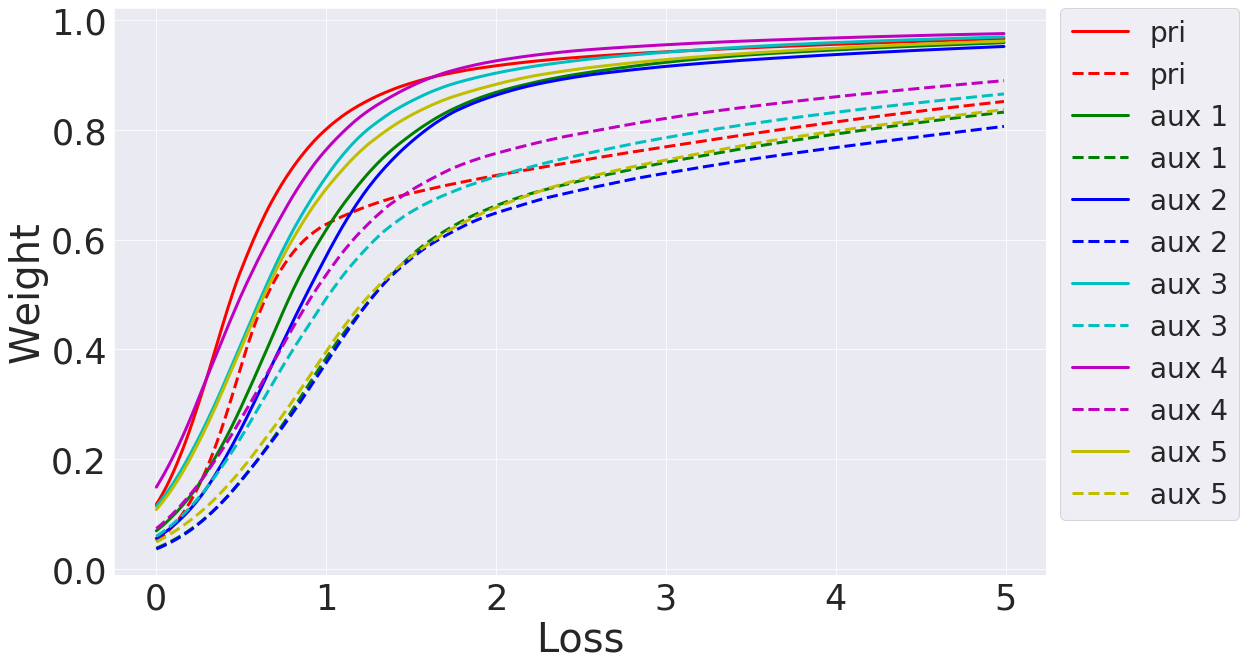}%
    	\includegraphics[height=3.7cm]{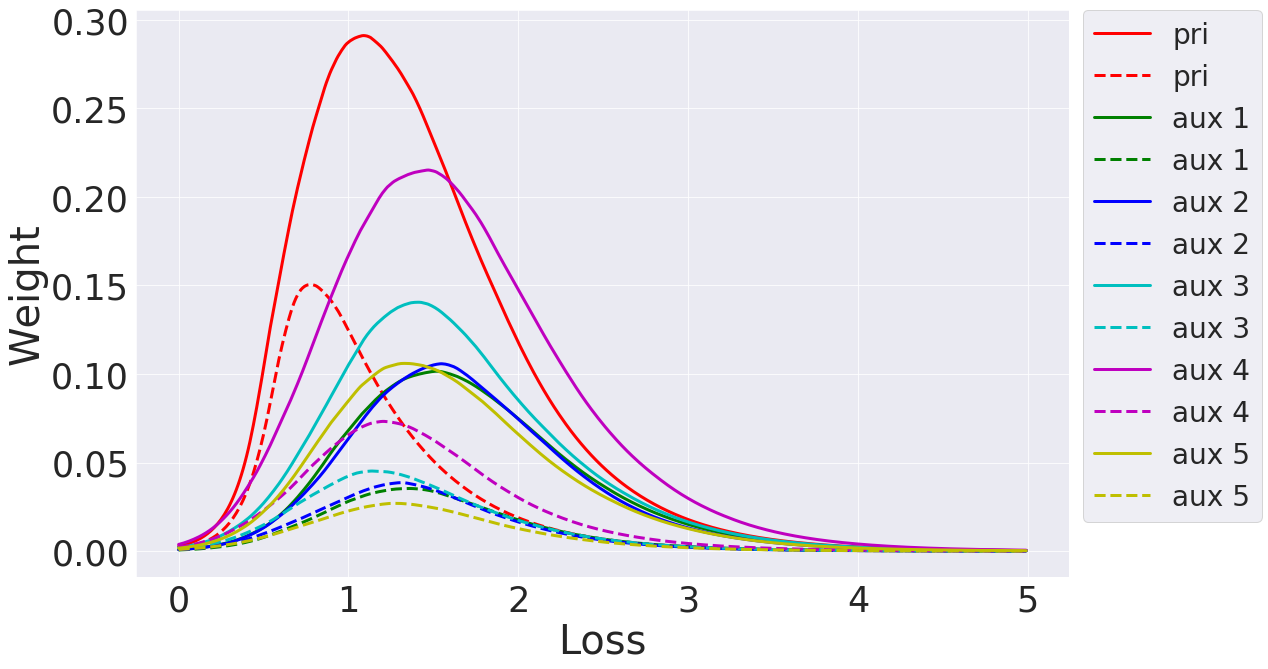}
    	}
\subfigure[GIN]{
    	\includegraphics[trim=0 0 210 0, clip,height=3.7cm]{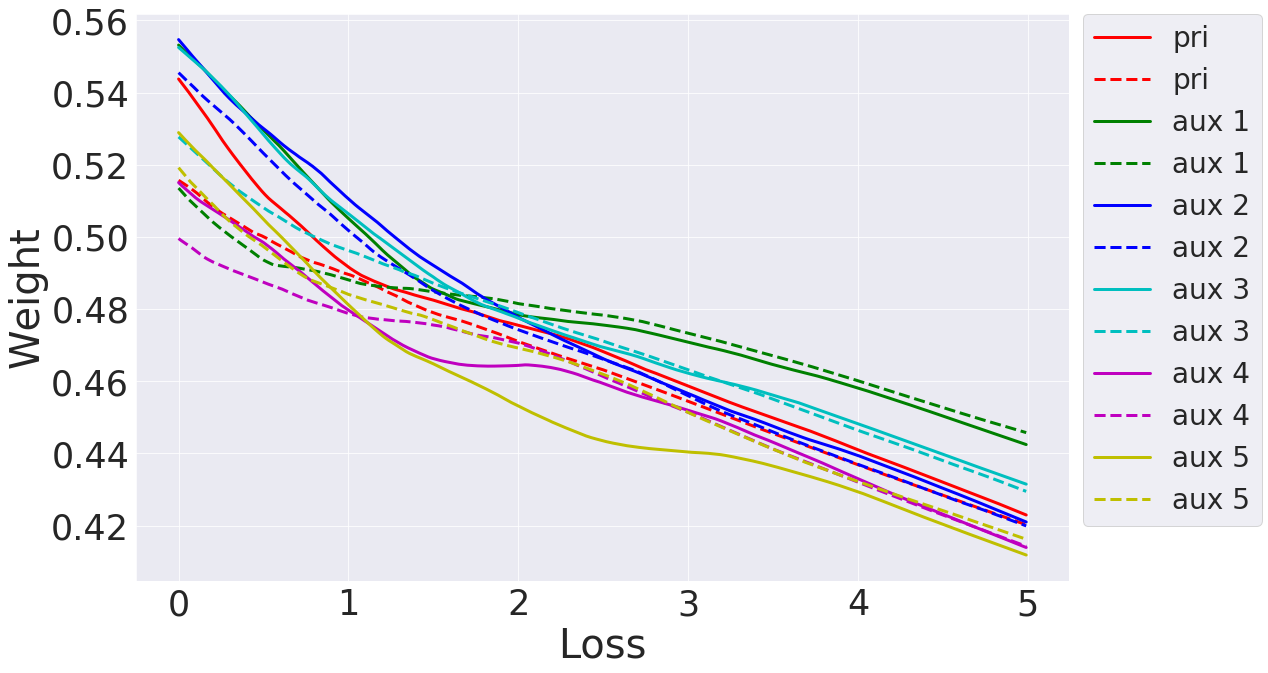}%
    	\includegraphics[trim=0 0 210 0, clip,height=3.7cm]{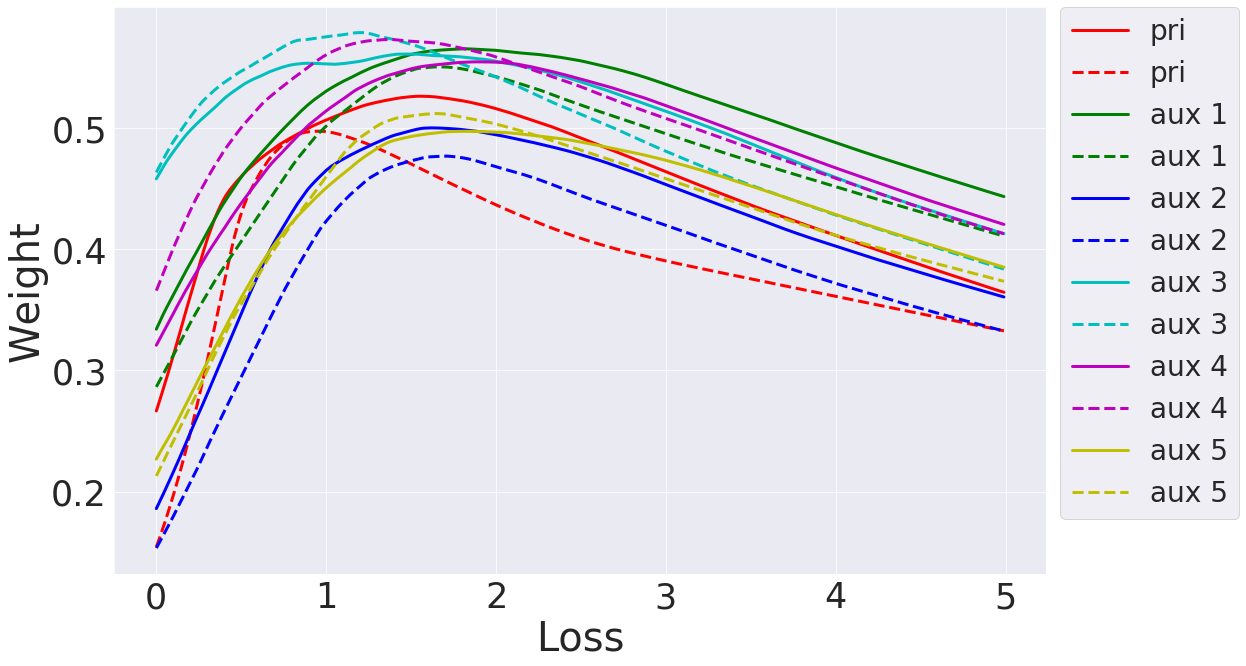}%
    	\includegraphics[height=3.7cm]{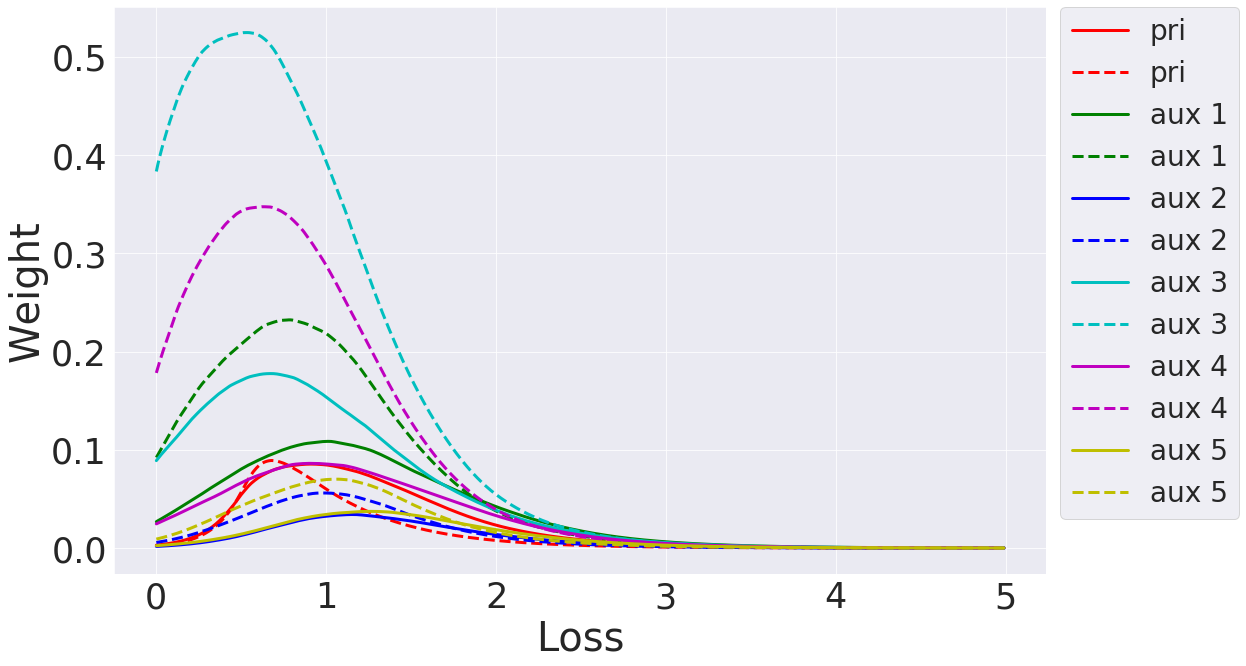}
    	}
\subfigure[SGC]{
    	\includegraphics[trim=0 0 210 0, clip,height=3.7cm]{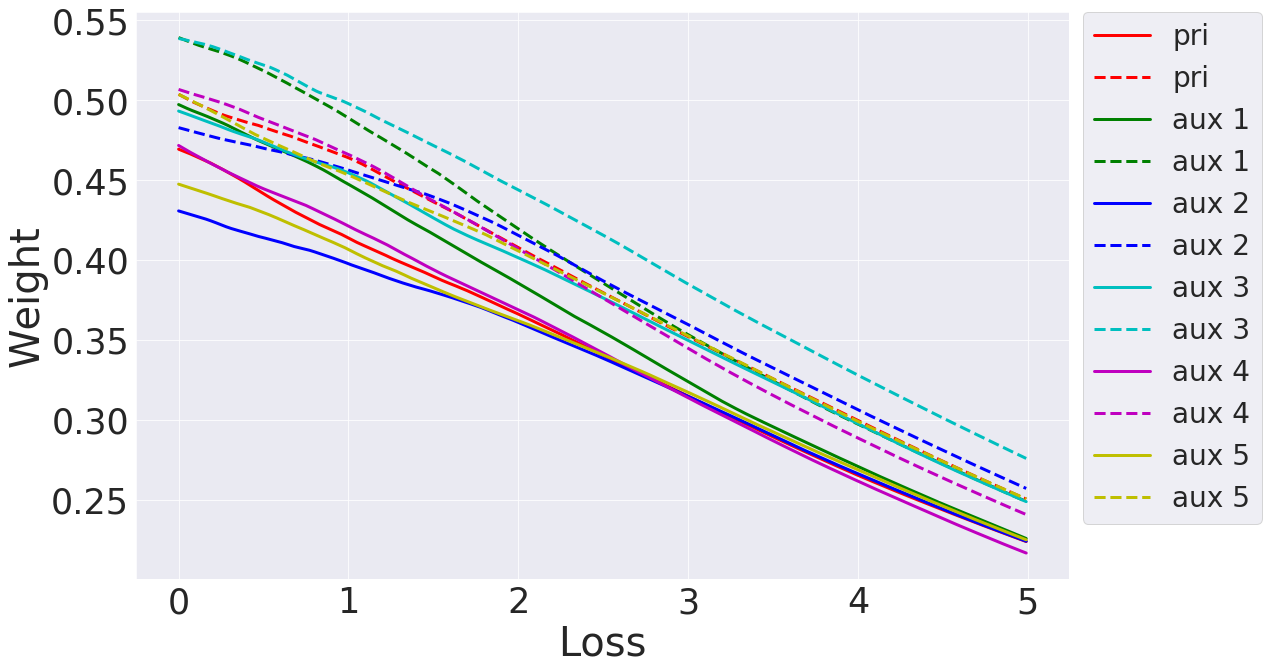}%
    	\includegraphics[trim=0 0 210 0, clip,height=3.7cm]{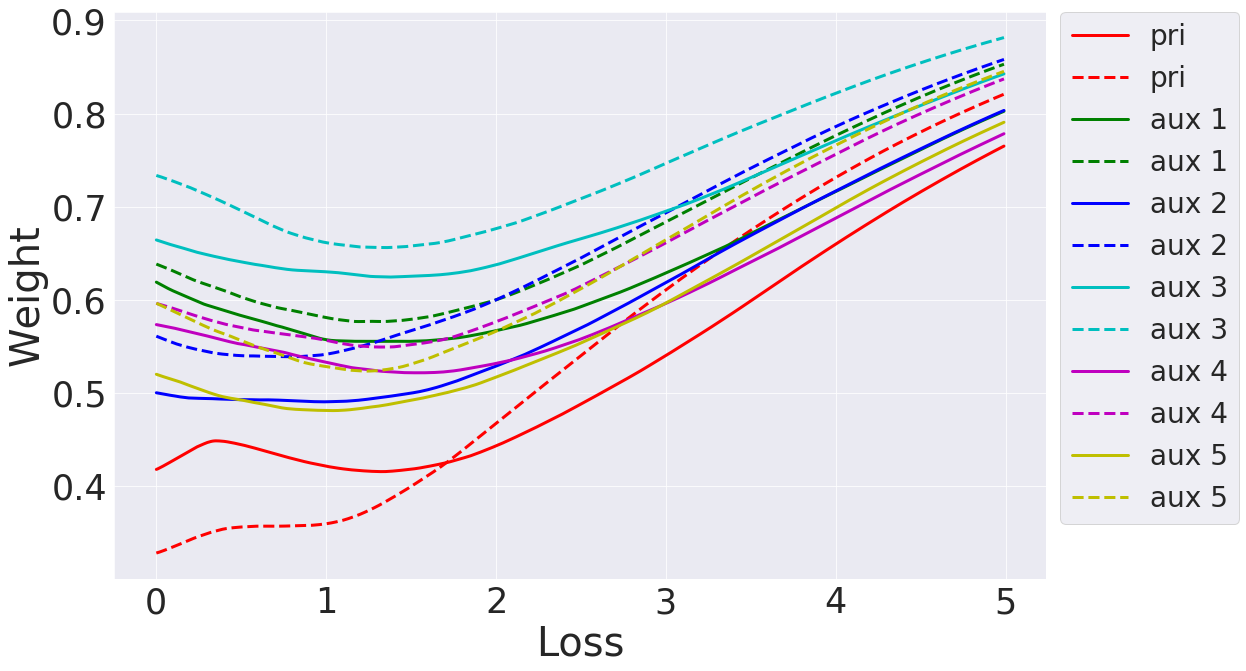}%
    	\includegraphics[height=3.7cm]{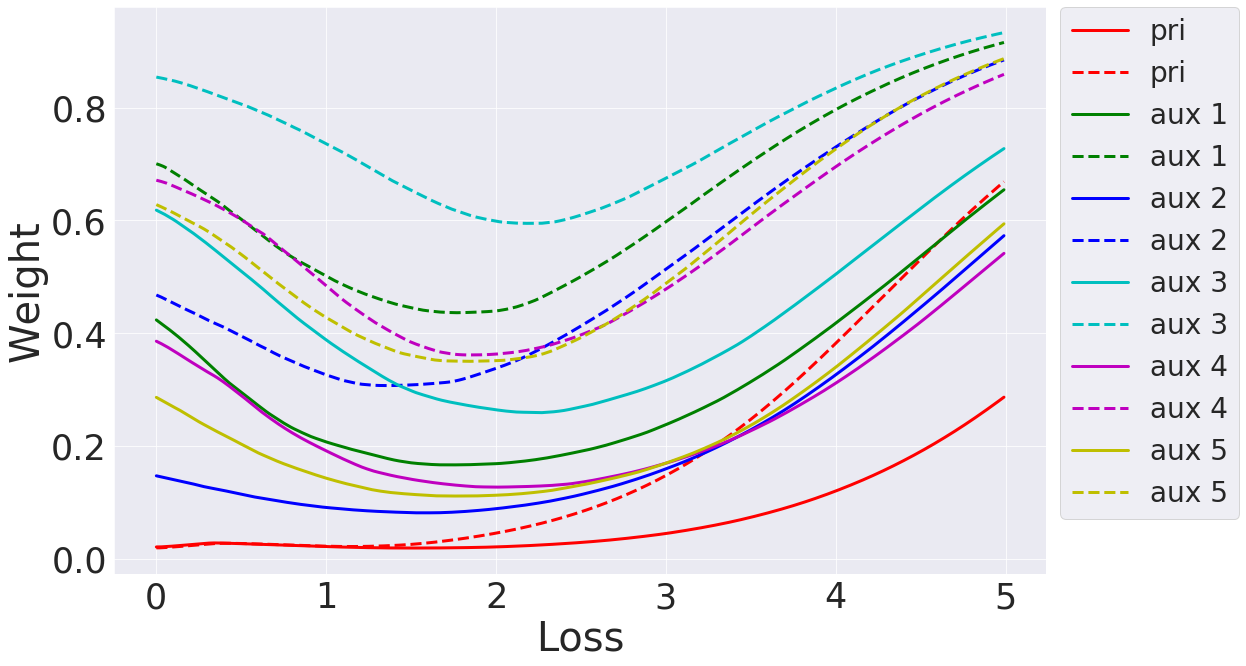}
    	}
\subfigure[GTN]{
    	\includegraphics[trim=0 0 210 0, clip,height=3.7cm]{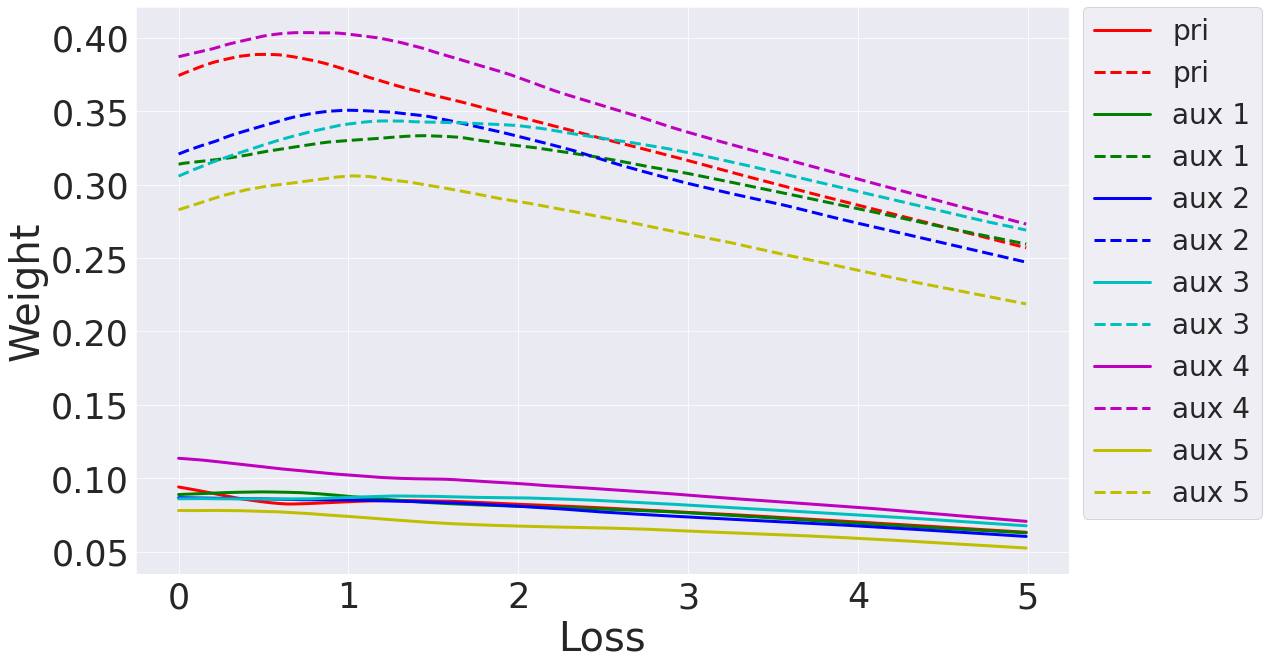}%
    	\includegraphics[trim=0 0 210 0, clip,height=3.7cm]{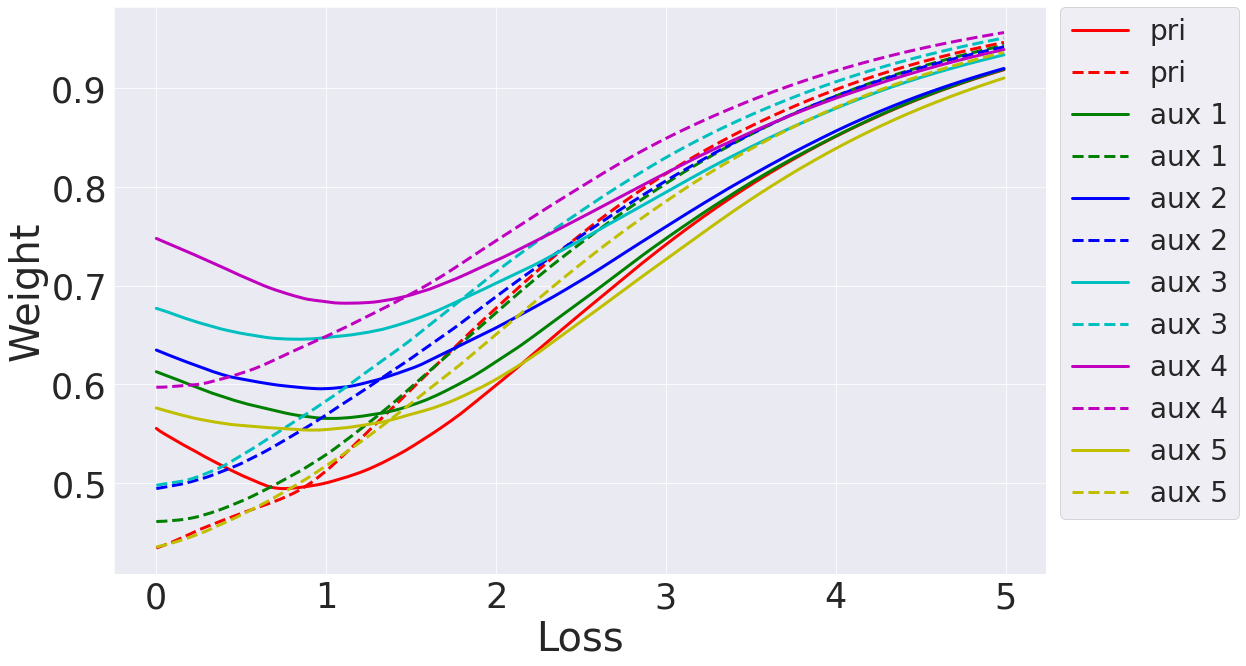}%
    	\includegraphics[height=3.7cm]{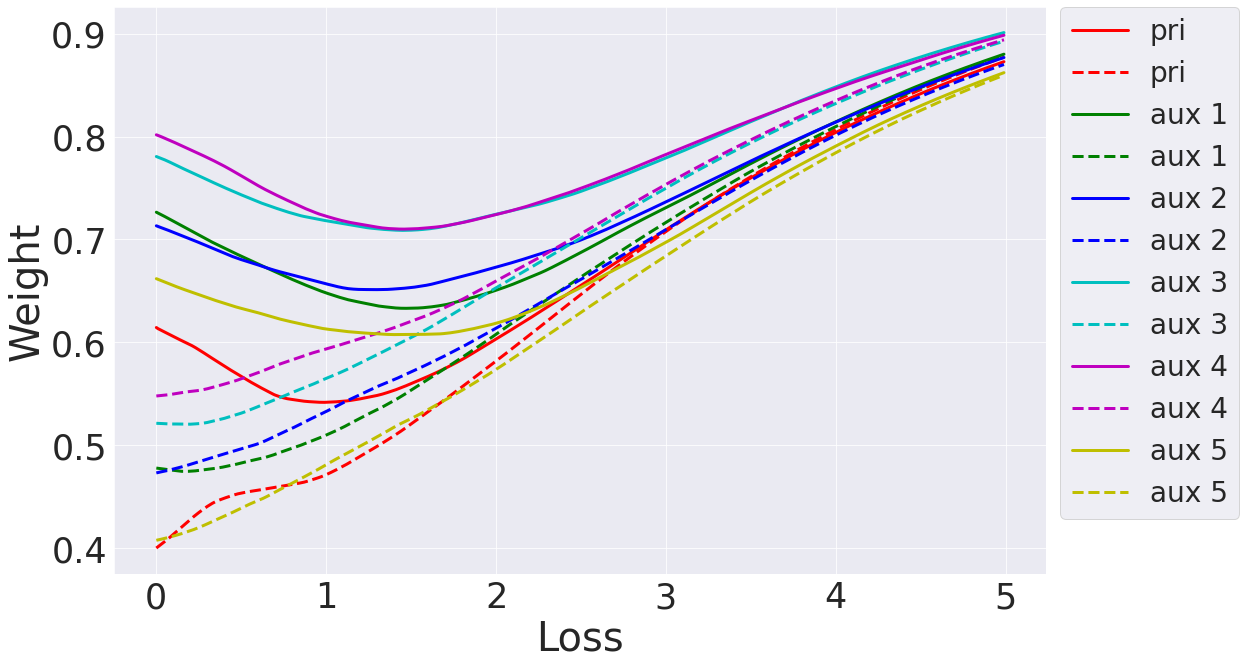}
    	}
\caption{\hjk{Weighting function $\Vc(\cdot)$ learnt by SELAR+Hint on Last-FM for each base model (GCN, GAT, GIN, SGCm and GTN). In each row from the left the first epoch, the best epoch with the best validation performance, and the last epoch are presented. The positive sample of the primary task is a red solid line  and the negative sample is a dashed line.
 }}
\label{fig:all_weights}
\end{figure*}
%%%%%%%%%%%%%%%%%%%%%%%%%%%%%%%%%%%%%%%%%%%%%
\subsection{Additional auxiliary tasks}
To demonstrate that our methods can be extended to include other auxiliary tasks, we experiment with multiple self-supervised tasks on a graph as auxiliary tasks. The details of auxiliary tasks we used are summarized below.

\textbf{Degree prediction} is to predict the degree for each node in the graph. The goal of this auxiliary task is to learn local structure information by quantifying the node connectivity as a local measure.
Specifically, we first calculate the degree $ d(i) = \sum_{j = 1}^{N} \mathbf{A}_{ij}$ for every node and group the nodes into three categories $c_i$; the top 20\%, bottom 20\% of the degree distribution, and the rest. 
We use randomly selected $|V_{\text{deg}}|$ samples from nodes $V$, \ie, $V_{\text{deg}} \subset V$. 
The loss of this auxiliary task can be formulated as the cross entropy loss as follows
\begin{align}
\Lc_{deg} = \frac{1}{|V_{\text{deg}}|} \sum_{i = 1}^{|V_{\text{deg}}|} \ell(f( x_i; \wb), c_i)
\label{degree}
\end{align}

\textbf{Pairwise Node Distance} \cite{jin2020self} is designed to predict the distance between a pair of nodes. 
It aims to encourage the GNNs to learn global topology information via a pairwise comparison. 
The shortest path length on the graph is used as the distance between nodes. 
The shortest path length $p_{ij}$ is pre-computed for all node pairs $(u_i, u_j)$ in our experiments. We group the node pairs in a graph into four categories and use only randomly selected $|E_{\text{dist}}|$ samples from all edges $E$, i.e., $E_{\text{dist}} \subset E$. The pairwise distance loss function is given as follows
\begin{align}
\Lc_{dist} = \frac{1}{|E_{\text{dist}}|} \sum_{(u_i, u_j) \in E_{\text{dist}}} \ell(|f( u_i; \wb)-f( u_j; \wb)|, c_{p_{ij}})
\end{align}

\textbf{PageRank prediction} aims to predict the PageRank score, which indicates the importance of each node in the graph. Roughly speaking, the assumption of the PageRank \cite{page1999pagerank} algorithm is that the more important nodes have more connections (or edges) with important nodes. 
% Predicting these PageRank scores makes models learn the importance of nodes based on the structural roles of nodes in the graph. 
$PR(i)$ = $\sum_{(j,i) \in E} \frac{PR(j)}{\text{out-degree}(j)}$ denotes the PageRank score of $i$-th node. 
Similar to degree prediction, we formulate it as a classification task and the cross-entropy is used as a loss function as in \eqref{degree}.

\textbf{Node clustering} first clusters nodes using the K-means clustering algorithm with node features and defines a self-supervised task that predicts which cluster a target node belongs to.
Given the node set $\mathcal{V}$, $\mathcal{A}_k \subset \mathcal{V}$ denotes a subset of the nodes belong to $k$-th cluster. 
The clustering algorithm will output $K$ disjoint subsets $\mathcal{A}_1, \dotsc, \mathcal{A}_K$. 
The label of an arbitrary node $u$ is the cluster index where the node $u$ belongs to, \ie, $y_u = k$ if $u \in \mathcal{A}_k$.
This task encourages models to learn similar features for nodes in the same cluster.

\textbf{Graph partitioning} is a self-supervised task that predicts which partition each node is assigned to.
We first separate nodes into $K$ disjoint subsets $\mathcal{P}_1, \dots, \mathcal{P}_K \subset \mathcal{V}$ using the METIS~\cite{karypis1998fast} algorithm. 
The partition algorithm generates partitions to minimize the number of connected edges (edge-cut) between partitions and the resulting partitions satisfy the following conditions: $\bigcup_{k=1}^{K} \mathcal{P}_k = \mathcal{V}$, $\forall k=1,\dotsc,K$ and $\mathcal{P}_i \cap \mathcal{P}_j = \emptyset$, $\forall i,j=1,\dotsc,K$ ($i \neq j$). 
The partition indices are used as self-supervised labels. 
It is similar to node clustering, but graph partitioning minimizes the partitioning cost (\ie, the number of removed edges to partition) rather than clustering nodes based on a similarity measure. 

To evaluate the benefits of our method in various combinations of the auxiliary tasks described above in node classification on IMDB, we compare our \textbf{SELAR} with standard multi-task learning (denoted as \textbf{MTL}).
We conducted experiments all base models based on six auxiliary tasks: \textbf{Degree}, node degree prediction; \textbf{Distance}, pairwise node distance prediction; \textbf{PageRank}, PageRank score prediction; \textbf{Clustering}, node clustering label prediction; \textbf{Partition}, graph partitioning label prediction; \textbf{Mata-path}, meta-path prediction proposed in Section~\ref{meta-path-prediction}.

Table~\ref{tab:multiple} shows that employing multiple auxiliary tasks could benefit base models in various settings. 
As expected, in combinations of multiple auxiliary tasks, SELAR consistently outperforms most base models trained by vanilla learning schemes only with the primary task (denoted as `\textbf{-}') as well as multi-task learning (denoted as \textbf{MTL}). 
For instance, SELAR provides more optimized auxiliary learning resulting in  2.8\% and 2.9\% improvement on SGC and GTN trained by pairwise node distance and has 3.2\% gains on GAT trained by PageRank prediction.
Particularly, SELAR provides 3.9\% improvement compared to MTL in trained GCN with degree prediction, pairwise node distance, PageRank prediction, and graph partitioning.
Interestingly, learning with various auxiliary task combinations, especially SELAR, on GTN and GCN improves 3.9\% and 3.0\% on average across all auxiliary task combinations compared to the vanilla learning scheme (only primary task). 
There are cases where auxiliary tasks interfere with the primary task and especially MTL often shows marginal degradation.
In comparison, SELAR show no degradation compared to Vanilla except for GIN cases.

\subsection{Meta cross-validation}
Meta cross-validation, i.e., cross-validation for meta-learning, helps to keep weighting function from  over-fitting on meta data. 
Table~\ref{tab:metacv_edge} evidence that our algorithms as other meta-learning methods can overfit to meta-data. 
As in Algorithm \ref{alg:selar}, our proposed methods, both SELAR and SELAR with HintNet,  with cross-validation denoted as `3-fold' alleviates the meta-overfitting problem and provides a significant performance gain, whereas without meta cross-validation denoted as `1-fold' the proposed method can underperform the vanilla training strategy.

%%%%%%%%%%%%%%%%%%%%%%%%%%%%%%%%%%%%%%%%%%%%%
\begin{table}[h]
  \caption{Comparison between 1-fold and 3-fold as meta-data on Last-FM.}
  \label{tab:metacv_edge}
  \footnotesize
  \centering
  \begin{tabular}{c|c|cc|cc}
    \toprule
     & & \multicolumn{2}{c}{SELAR} & \multicolumn{2}{c}{SELAR+Hint}\\
     
     Model &Vanilla &1-fold & 3-fold & 1-fold & 3-fold  \\
    \midrule
     GCN & 0.7963 & 0.7885 & \textbf{0.8296} & 0.7834 & \textbf{0.8121}  \\
     GAT & 0.8115 & 0.8287 & \textbf{0.8294} & 0.8290 & \textbf{0.8302}  \\
     GIN & 0.8199 & 0.8234 & \textbf{0.8361} & 0.8244 & \textbf{0.8350}  \\
     SGC & 0.7703 & 0.7691 & \textbf{0.7827} & 0.7702 & \textbf{0.7975}  \\
     GTN & 0.7836 & 0.7897 & \textbf{0.7988} & 0.7915 & \textbf{0.8067}  \\
    \bottomrule
  \end{tabular}
\end{table}
% \vspace{-1cm}
%%%%%%%%%%%%%%%%%%%%%%%%%%%%%%%%%%%%%%%%%%%%%
\section{Analyses}
The effectiveness of meta-path prediction and the proposed learning strategies are answered above.
To address the last research question \textbf{Q4.} why the proposed method is effective, we provide analysis on the weighting function $\Vc(\xi;\Theta)$ learned by our framework. Also, we show the evidence that meta-overfitting occurs and can be addressed by cross-validation as in Algorithm \ref{alg:selar}.

\subsection{Weighting Function}
Our proposed methods can automatically balance multiple auxiliary tasks to improve the primary task. 
To understand the ability of our method, we analyze the weighting function and the adjusted loss function by the weighting function, i.e.,$\Vc(\xi;\Theta)$, $\Vc(\xi;\Theta)\ell^t(y,\hat{y})$. 
The positive and negative samples are solid and dash lines respectively.
We present the weighting function learnt by SELAR+HintNet for GAT which is the best-performing construction on Last-FM. The weighting function is from the epoch with the best validation performance.
Fig.~\ref{fig:weighfunc} shows that the learnt weighting function attends to hard examples more than easy ones with a small loss range from 0 to 1.

Also, the primary task-positive samples are relatively less down weighted than auxiliary tasks even when the samples are easy (i.e., the loss is ranged from 0 to 1). Our adjusted loss $\Vc(\xi;\Theta)\ell^t(y,\hat{y})$ is closely related to the focal loss, $-(1-p_t)^{\gamma}\log(p_t)$. When $\ell^t$ is the cross-entropy, it becomes $\Vc(\xi;\Theta)\log(p_t)$, where $p$ is the model's prediction for the correct class and $p_t$ is defined as $p$ if $y=1$, otherwise $1-p$ as \cite{lin2017focal}.
The weighting function differentially evolves over iterations. At the early stage of training, it often focuses on easy examples first and then changes its focus over time. Also, the adjusted loss values by the weighting function learnt by our method differ across tasks. To analyze the contribution of each task, we calculate the average of the task-specific weighted loss on the Last-FM and Book-Crossing datasets. Especially, on the Book-Crossing, our method has more attention to 'user-item' (primary task) and `user-item-literary.series.item-user' (auxiliary task) which is a meta-path that connects users who like a book series. This implies that two users who like a book series likely have a similar preference.

\subsection{Weighting function at different training stages}
The weighting functions of our methods dynamically change over time. 
In Fig. \ref{fig:all_weights}, each row is the weighting function learned by SELAR+HintNet for GCN~\cite{GCN}, GAT~\cite{GAT}, GIN~\cite{xu2018powerful}, and SGC~\cite{wu2019simplifying} in order on Last-FM. 
From left, columns are from the first epoch, the epoch with the best validation performance, and the last epoch respectively. 
The positive and negative samples are illustrated in solid and dash lines respectively in Fig.~\ref{fig:all_weights}. 
At the begging of training (the first epoch), one noticeable pattern is that the weighting function focuses more on `easy' samples.
At the epoch with the highest performance, easy samples are down-weighted and the weight is large when the loss is large. 
It implies that hard examples are more focused.
At the last epoch, most weights converge to zero when the loss is extremely small or large in the last epoch. 
Since learning has almost completed, the weighting function becomes relatively smaller in most cases, e.g., GCN, GAT, and GIN.
Especially, for GCN and GAT in the epoch with the highest performance, the weights are increasing and it means that our weighting function imposes that easy samples to smaller importance and more attention on hard samples. 
Among all tasks, the scale of weights in the primary task is relatively high compared to that of auxiliary tasks. 
This indicates that our method focuses more on the primary task.
\label{sec:exp_analysis}

\section{Conclusion}
\label{sec:conclusion}

We proposed a framework that learns to softly select auxiliary tasks to improve the primary task.
The auxiliary tasks can be further improved by our proposed method SELAR, which automatically balances auxiliary tasks to assist the primary task via a form of meta-learning.
Besides, we introduced meta-path prediction as self-supervised auxiliary tasks on heterogeneous graphs. 
Our experiments show that the representation learning on heterogeneous graphs can benefit from meta-path prediction which encourages capturing rich semantic information.
The learnt weighting function identifies more beneficial auxiliary tasks (meta-paths) for the primary tasks. 
Within a task, the weighting function can adjust the cross entropy like the focal loss, which focuses on hard examples by decreasing weights for easy samples. 

Moreover, when it comes to challenging and remotely relevant auxiliary tasks, 
our HintNet helps the learner by correcting the learner's answer dynamically and further improves the gain from auxiliary tasks.
Our framework based on meta-learning provides learning strategies to balance the primary task with different auxiliary tasks.
Further, our method identifies easy/hard (and positive/negative) samples between tasks or within each task.
We demonstrated the effectiveness of our proposed methods to improve the representational power of GNNs in node classification, link prediction, and recommender systems.
Interesting future directions include applying our framework to other domains such as computer vision, natural language processing.

\ifCLASSOPTIONcaptionsoff
  \newpage
\fi

\bibliographystyle{unsrt}
\bibliography{Ref}

\end{document}